%% file: main.tex
\documentclass[11pt]{article}

\usepackage[final]{acl}
\usepackage{booktabs}
\usepackage{subcaption}
\usepackage{times}
\usepackage{latexsym}
\usepackage{enumitem}
\usepackage{todonotes}

\usepackage[T1]{fontenc}

\usepackage{amsfonts}
\usepackage{amsmath}
\usepackage[utf8]{inputenc}
\usepackage{tcolorbox}
\tcbuselibrary{listings, breakable}
\usepackage{listings}
\usepackage{multirow}
\usepackage{microtype}

\usepackage{inconsolata}

\usepackage{graphicx}
\usepackage{xcolor}

\definecolor{darkred}{RGB}{150,40,40}
\definecolor{improveblue}{RGB}{0,70,140}

\usepackage{tcolorbox}
\usepackage{booktabs}
\usepackage{array}
\usepackage{xcolor}
\usepackage{enumitem}

\tcbset{
  aibox/.style={
    top=8pt,
    bottom=3pt,
    colback=white,
    colframe=black,
    colbacktitle=black,
    enhanced,
    center,
    attach boxed title to top left={yshift=-0.1in,xshift=0.15in},
    boxed title style={boxrule=0pt,colframe=white,},
  }
}
\newtcolorbox{AIbox}[3][]{aibox, width=#2, title=#3,#1}

\tcbuselibrary{skins, breakable}

\definecolor{tomcolor}{RGB}{70, 130, 180}
\definecolor{johncolor}{RGB}{60, 179, 113}
\definecolor{bggray}{RGB}{248, 248, 248}
\definecolor{deltabg}{RGB}{240, 240, 245}

\makeatletter
\newcommand{\dashedrule}{%
  \noalign{\vskip 0.5ex}%
  \multispan{8}\leaders\hbox to 3pt{\hss.\hss}\hfill\kern0pt\\[-0.8ex]%
  \noalign{\vskip 0.5ex}%
}
\makeatother

\title{Sentipolis: Emotion-Aware Agents for Social Simulations}

\author{
 \textbf{Chiyuan Fu\textsuperscript{1}\thanks{Equal Contributions}},
 \textbf{Lyuhao Chen\textsuperscript{1}\footnotemark[1]},
 \textbf{Yunze Xiao\textsuperscript{1}\footnotemark[1]},
 \textbf{Weihao Xuan\textsuperscript{2,3}},
 \textbf{Carlos Busso\textsuperscript{1},
 \textbf{Mona Diab\textsuperscript{1}}
 }
\\
\\
 \textsuperscript{1}Carnegie Mellon University,
 \textsuperscript{2}The University of Tokyo,
 \textsuperscript{3}RIKEN AIP\\
 \texttt{chiyuanf@alumni.cmu.edu \{lyuhaoc,yunzex,cbusso,mdiab\}@andrew.cmu.edu}\\
 \texttt{xuan@ms.k.u-tokyo.ac.jp}
}

\begin{document}
\maketitle

\input{main/0-abstract}
\input{main/1-introduction}
\input{main/2-related_works}
\input{main/3-method}
\input{main/4-experiment}
\input{main/5_sotopia}
\input{main/6-network}

\input{main/7-conclusion}

\input{main/limitations}

\bibliography{custom,anthology_0}

\input{appendix/main}


\end{document}

%% file: main/0-abstract.tex
\begin{abstract}
LLM agents are increasingly used for social simulation, yet emotion is often treated as a transient cue, causing \emph{emotional amnesia} and weak long-horizon continuity. We present \textsc{Sentipolis}, a framework for emotionally stateful agents that integrates continuous Pleasure-Arousal-Dominance (PAD) representation, dual-speed emotion dynamics, and emotion--memory coupling. Across thousands of interactions over multiple base models and evaluators, \textsc{Sentipolis} improves emotionally grounded behavior, boosting communication, and emotional continuity. Gains are model-dependent: believability increases for higher-capacity models but can drop for smaller ones, and emotion-awareness can mildly reduce adherence to social norms, reflecting a human-like tension between emotion-driven behavior and rule compliance in social simulation. Network-level diagnostics show reciprocal, moderately clustered, and temporally stable relationship structures, supporting the study of cumulative social dynamics such as alliance formation and gradual relationship change.
\end{abstract}

%% file: main/1-introduction.tex
\section{Introduction}

Recent advances in reasoning \cite{li202512surveyreasoning} and long-context memory \cite{hu2025memoryageaiagents} are making large language models (LLM) appear increasingly human-like \cite{xiao-etal-2025-humanizing}, which has led researchers to adopt LLM agents as a substrate for social simulation \cite{anthis2025llmsocialsimulationspromising,Agrawal_Xiao_2026}. Recent studies have focused on a wide range of applications, such as education \cite{zhang-etal-2025-simulating, yuan2026validstudentsimulationlarge}, public policy \cite{hou2025societygenerativeagentssimulate,li2025policyuseful}, social dynamics \cite{park2023generative}, and debates \cite{liu-etal-2025-synthetic}. These efforts have shown the potential of scalable simulations, but they have also highlighted the open challenges in long-running interactive simulations, including grounding, calibration, and behavioral validity \cite{li2025llmgeneratedpersonapromise}.

\begin{figure}[t]
  \includegraphics[width=\columnwidth]{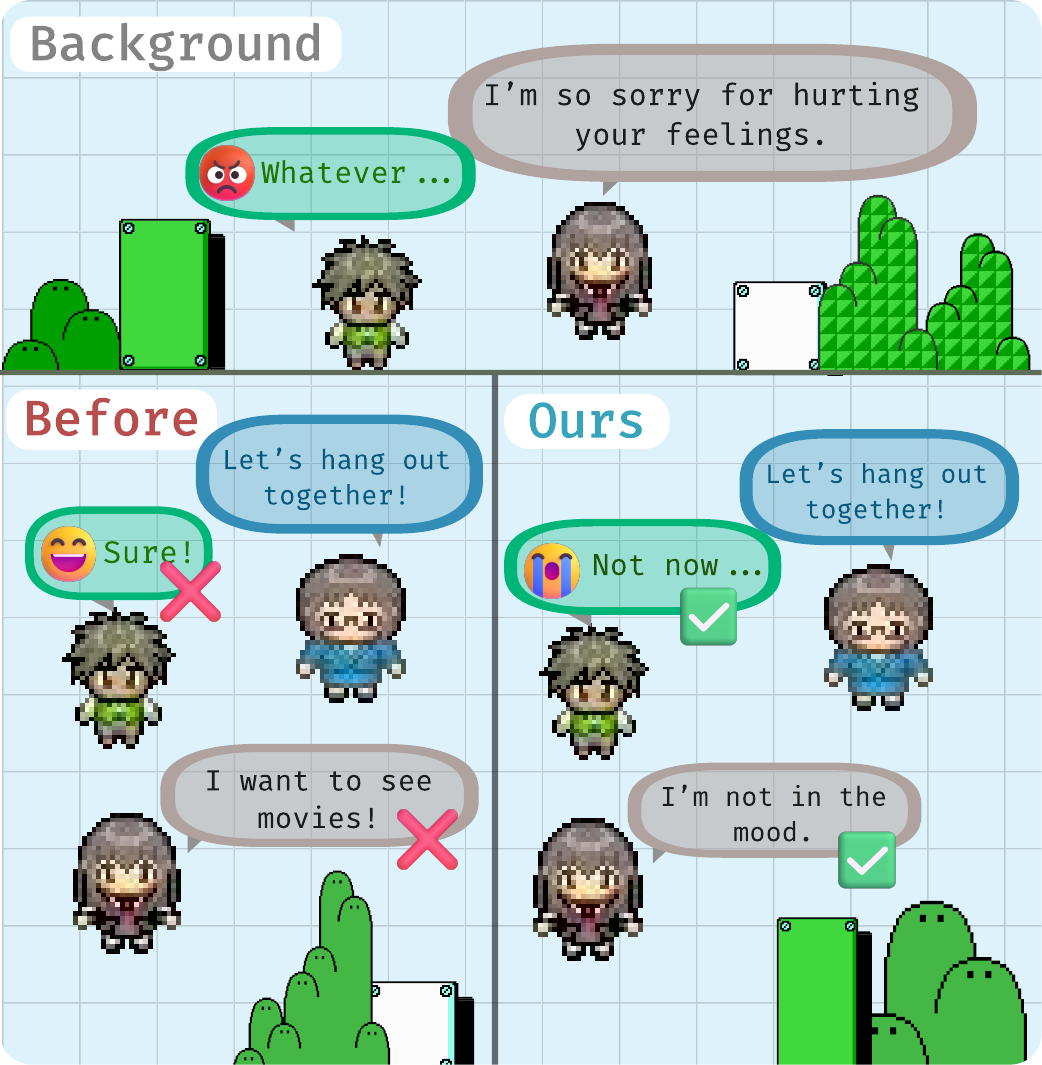}
\caption{An example of emotional amnesia in LLM-based social simulations. Bob and Alice had an argument and they carried a negative emotion. Without persistent emotion modeling, agents lead to emotionally inconsistent responses, whereas emotion-aware agents preserve emotional continuity and produce responses consistent with their history.}
\end{figure}

However, \textbf{emotion-aware mechanisms in existing LLM social simulations are rarely designed for long-horizon emotional continuity}. Prior work has either completely omitted explicit emotion state \citep{park2023generative}, treated emotion as a short-horizon signal \citep{regan2024generativeagentspredictemotion}, or introduced Pleasure-Arousal-Dominance (PAD) variables with hand-designed update rules \citep{ma2025emotionalignment}. These approaches overlook that human social interaction is emotionally stateful: emotions evolve during conversations \cite{goodwin2000emotion}, carry over across encounters \cite{kuppens2010emotional}, and shape subsequent interpretation and response \cite{schwarz1983mood}. When agents lack a persistent emotional state, being insulted may not increase irritability in later turns, and repeated positive exchanges may not accumulate into stronger bonds. We refer to this failure mode as \textbf{emotional amnesia}.

\begin{figure*}[!t]
  \includegraphics[width=\textwidth]{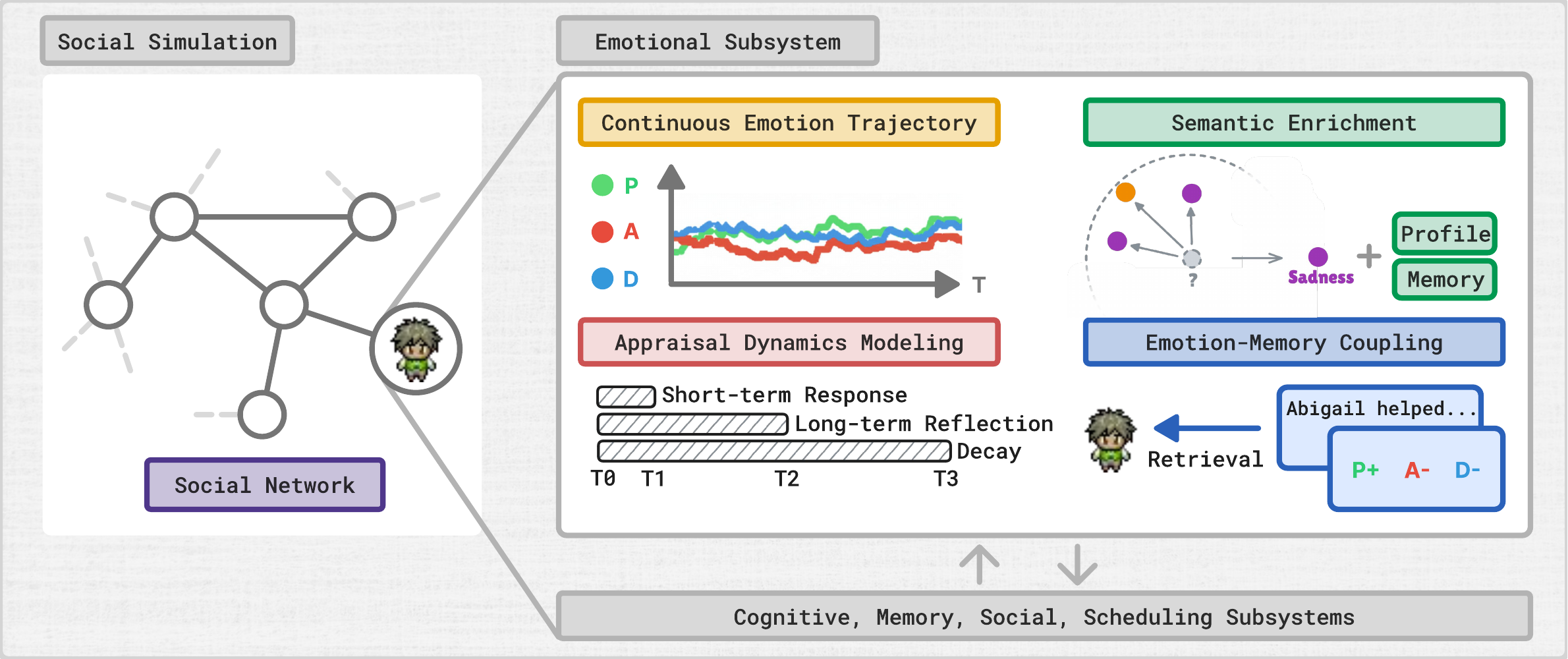}
  \caption{Overview of our simulation setup. Multiple agents form a social network, where each agent is an orchestration of different subsystems. In an emotional subsystem, we integrate explicit emotion modeling (continuous emotion trajectory, appraisal dynamics) with implicit emotion modeling (emotional memory and semantic enrichment) to achieve a balance between controllability and realism.}
  \label{fig:overview}
\end{figure*}

Motivated by \textbf{emotional amnesia}, we treat emotion as a first-class persistent state at the agent architecture level. Crucially, we do not claim that the base LLM internalizes affect in its parameters; instead, we maintain emotion as an explicit, interpretable module that is backbone-agnostic and can be layered onto any LLM agent framework. This design prioritizes controllability, interpretability, and modular deployment over end-to-end integration. Concretely, each agent maintains a continuous PAD vector that evolves throughout the simulation. This persistent state further supports a dual-speed emotion dynamic model with decay, aligning with the perspective of the Emotion and Adaptation model \cite{MARSELLA200970}. To make emotion consequential for future behavior, we couple emotion to memory by storing events and reflections together with their PAD-derived emotion tags. For interpretability and prompt control, we map each continuous PAD state to a semantic emotion description via KNN over human PAD anchor points, and then generate a vivid emotion paragraph conditioned on the label, the agent profile, and retrieved memories; this paragraph is injected into the prompt to ground dialog and reflection in the agent's current emotion state.

Across different models and evaluators, emotional statefulness yields clear gains in both emotional intelligence and social competence, as measured with communication metric improving by about 30\% on average and continuity metrics roughly doubling. Component-level ablations confirm that these gains are not reducible to generic prompt structuring: different modules serve distinct functional roles, with PAD coupling and open-vocabulary description primarily supporting continuity, while decay and KNN retrieval primarily support believability and communication quality. Human annotators show strong inter-annotator agreement (Krippendorff's $\alpha$ up to 0.871) and substantial agreement with the LLM judge ($\alpha = 0.825$ pooled), validating the evaluation protocol. The effects are heterogeneous: believability increases for higher-capacity models (GPT-5.2, Grok-4) but drops for the smaller model. Empathy improves for GPT-4o-mini and GPT-5.2, slightly declines for Grok-4. Social-rule scores are often unchanged and, when they shift, become only mildly more negative under some evaluators rather than showing systematic penalties. Network-level diagnostics further indicate that emotion-memory coupling produces high reciprocity, moderate clustering, and stable structure over time, supporting the study of cumulative social dynamics such as alliance formation and gradual relationship change.

Our contribution is as follows:

\begin{itemize}[nosep, topsep=0pt]
    \item We identify long-horizon emotional continuity as a missing design target in LLM agent simulations and characterize \emph{emotional amnesia} as a concrete failure mode of missing emotion carryover;
    \item We propose an emotionally stateful agent architecture that integrates PAD, \emph{dual-speed emotion dynamics}, and emotion-tagged memory to support emotion carryover across encounters. To the best of our knowledge, this is the first work within LLM-based social simulation to introduce a dedicated persistent emotion system with PAD dynamics and emotion-aware memory interaction and to evaluate its long-horizon behavioral effects;
    \item We validate the architecture through component-level ablations that reveal structured, non-uniform contributions of individual modules, and through human evaluation that confirms strong agreement with the LLM-judge protocol;
    \item We show that emotional statefulness yields more realistic changes within-agents in social intelligence on SotopiaEval \citep{zhou2024sotopiainteractiveevaluationsocial} compared to non-stateful baselines;
    \item We show that emotion-aware agents spontaneously develop reciprocal relationships and a stable community structure that is unseen in Generative Agent \cite{park2023generative}, providing network-level validation of the proposed architecture.
\end{itemize}

%% file: main/2-related_works.tex
\section{Related Work}
\subsection{LLM Agents for Social Simulation}
LLM agents in social simulation studies often adopt a modular design. To achieve long-term coherence in simulation, Generative Agents \citep{park2023generative} designed a retrieval-based memory stream architecture, which was refined in later work via learned retrieval networks \cite{hong2025enhancing}. 
Beyond individual agents, agent groups exhibit emergent social dynamics such as spontaneous convention formation  \cite{dai2024artificialleviathanexploringsocial} and  group-level biases \cite{liu-etal-2025-synthetic}. They have been validated against established paradigms such as public-goods cooperation games, where simulated agents reproduce classic human patterns \cite{Piedrahita}. 
At larger scales, platforms like AgentSociety run thousands of agents with distinct profiles in open-ended worlds, reporting macro-level patterns that align with real diffusion and policy-response phenomena \cite{piao2025}. However, long-term grounding still breaks down in extended multi-session dialogs, where even retrieval-augmented models lag behind humans \cite{maharana-etal-2024-evaluating} and frequently misattribute events  \cite{ran2025bookworldnovelsinteractiveagent} or fail at temporal reasoning \cite{chen-etal-2025-perceive}, underscoring the need for stronger memory architectures and continual learning.

\subsection{Affective Computing in Conversational Agents}
Affective computing models emotion in dialog agents using either categorical representations (discrete classes \cite{lim-cheong-2024-integrating}) or dimensional representations (continuous attributees such as pleasure/valence, arousal, and dominance). Dimensional representations better capture subtle emotional blends, mapping emotion into the PAD space to determine the overall emotional state \cite{dong2025simulatinghumanbehaviorpsychologicalmechanism}.
Incorporating internal emotion variables improves social realism. For example, user valence correlates with task success \cite{feng-etal-2024-infusing}. Therefore, recent systems have added emotion recognition and management components that infer user feelings and adjust responses accordingly. They are often inspired by the appraisal theories such as the Ortony, Clore and Colling (OCC) model to link dialog events with emotion elicitation \cite{feng-etal-2024-infusing}. 
Emerging work suggests that endowing agents with "self-emotion" produces more human-like behavior. Agents with background emotional contexts use more varied dialog strategies, and roughly half of agents' decisions in multi-turn discussions change when their self-emotion is switched \cite{zhang-etal-2024-self-emotion}. The field is, thus, moving from simple sentiment tagging toward deeper integration of emotion as a behavioral driver, whether through dimensional emotional variables or cognitively motivated appraisal mechanisms.

\subsection{Emotional and Social Intelligence Evaluation in LLMs}
A growing line of work evaluates emotional intelligence and social intelligence in LLMs using standardized benchmarks. On the emotional intelligence side, prior benchmarks have assessed emotion recognition and empathetic response generation under short prompted scenarios, ranging from empathetic conversational settings \cite{rashkin-etal-2019-towards} to emotion-centric benchmark \cite{chen-etal-2024-emotionqueen} and situation-to-response tests  \cite{sabour-etal-2024-emobench}. Some have explicitly compared model reactions against human emotional judgments in realistic situations \cite{huang2024apathetic}. 
Other researchers have also tested LLMs on psychometric-style emotional intelligence items, suggesting that strong models can solve many standard emotional intelligence questions \cite{Schlegel2025-EItests}. On the social reasoning side, most benchmarks measure social intelligence as common sense reasoning about social interactions and norm \cite{sap-etal-2019-social} or as theory-of-mind competence \cite{wu-etal-2023-hi,Strachan2024-ToM}. However, most benchmarks only test short, one-off scenarios and treat emotion as an output, leaving it unclear whether social competence persists in long-horizon simulations or how explicit emotion dynamics shape behavior over time.

%% file: main/3-method.tex
\section{Sentipolis}
\label{sec:system-design}

\subsection{Simulation Framework and Agent Architecture}


\label{sec:sim-framework}
Following the Generative Agent framework\cite{park2023generative}, our system simulates a small sandbox world with 25 predefined agents that cover a diverse set of personas. In this simulation, agents observe events in different areas of the map, make plans, reflect on their goals, and trigger interactions with other agents, which is the primary driver of social dynamics in the simulation. 

Every agent is implemented as an orchestrated pipeline of modular subsystems. At each time step, the agent coordinates: (1) a cognition module that selects high-level intents and plans; (2) a memory module that retrieves relevant past events and summarizes them into contextual cues; (3) a movement module that updates locations and determines feasible encounters; (4) a social module that detects and instantiates possible new conversations; (5) a scheduler that manages time and aligns the agent's schedule with its current state; and, (6) an emotion module that updates emotion based on conversational events and longer-term reflection. The state of the module is also fed back to influence the inner dynamics of the cognitive, memory, and social subsystems.

\subsection{Emotion Representation}
\label{sec:emotion-mechanism}

The PAD emotional state model is a psychological model that describes and measures emotional states \cite{Mehrabian1996}. Each agent maintains a continuous emotion state in the Pleasure-Arousal-Dominance (PAD) space. The Pleasure-Displeasure Scale measures how pleasant or unpleasant one feels about something. The Arousal-Nonarousal Scale measures how energized or soporific one feels. The Dominance-Submissiveness Scale represents  how controlling versus controlled one feels \cite{mehrabian1980basic}. We treat PAD as a persistent internal state that can carry across encounters, enabling the simulator to represent both momentary reactions and longer-horizon mood-like tendencies.

To make emotion behaviorally consequential over long horizons, we couple emotion to memory at the representation level. When an agent creates a new memory, the accompanying emotional impact is also recorded, which can be retrieved later to condition emotional reasoning as well as downstream action selection.

\subsection{Appraisal Modeling}
Inspired by theoretical distinction between appraisal and inference in Emotion and Adaptation (EMA) \cite{MARSELLA200970}, we implement emotion updates at two coupled time scales. In appraisal theory, emotions arise from how an agent interprets its relationship to the environment, whether events are relevant to its goals, who caused them, and what coping resources are available. EMA argues that appraisal itself is a fast, automatic process; what varies in speed is the inference that constructs and updates the mental representation being appraised. Fast inference operates on immediately available information -- pattern recognition, retrieving associations, processing the current input. Slow inference integrates over broader context, draws on memory, and reasons about causes and implications. This maps naturally onto our two-timescale design: fast updates perform inference at the granularity of conversational turns, showing the immediate impact of current input; while slow updates perform inference during reflection, integrating retrieved history and accumulated experience. Further details are discussed in appendix \ref{appraisal_modeling_appendix}.

Emotions fade with time, and the duration of emotional experience is highly variable \cite{verduyn2011relation}. To model this adaptive phenomenon in human emotional experience, we designed an emotion decay mechanism. Let the PAD state be a vector $\mathbf{s}(t) = [P(t), A(t), D(t)]^\top$. We apply a half-life T1/2 decay at every time step.
$$
\mathbf{s}(t+\Delta t) \;=\; \mathbf{s}(t)\, 2^{-\Delta t / T_{1/2}}.
$$

\begin{figure}[t]
  \includegraphics[width=\columnwidth]{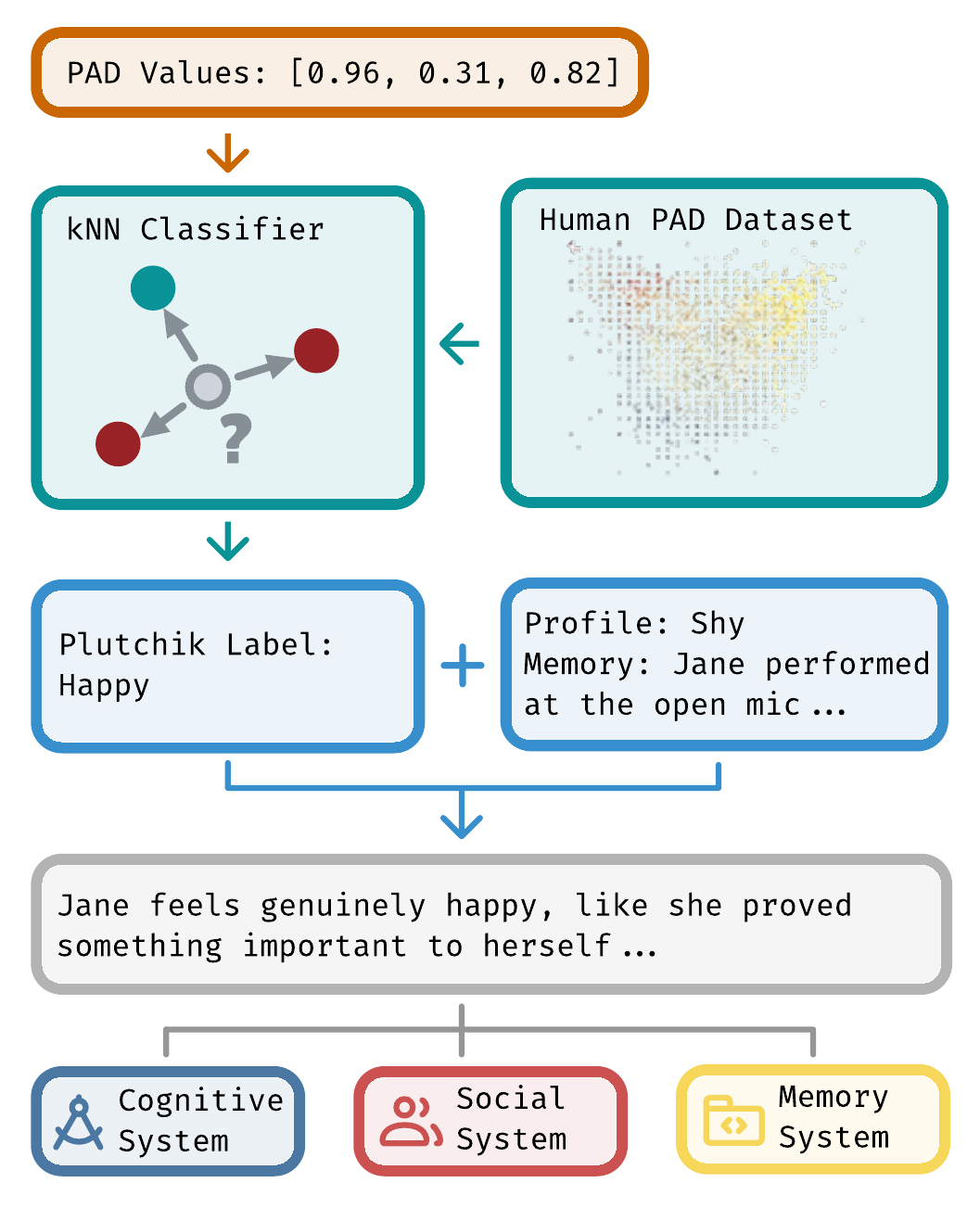}
  \caption{A simplified example of semantic enrichment. Instead of directly injecting emotion state into LLM's generation process, we designed a transformation pipeline which incorporates agent's memory, profile and real human emotion distribution data. }
  \label{fig:PAD}
\end{figure}

\subsection{Semantic Enrichment}
\label{sec:pad-language-transformation}

Downstream LLM prompts the benefits of human-interpretable emotion cues \cite{Li2023EmotionPrompt}, so we translate the agent’s continuous PAD state into a semantically meaningful emotion label and description. Concretely, we first perform k-Nearest Neighbor (KNN) on real human PAD data points \cite{busso2025msppodcastcorpus} to help the model interpret the semantic meaning of PAD values. Given the agent’s current PAD coordinate, we retrieve the nearest human reference points and map the state to a Plutchik-style emotional label \cite{Plutchik1980Emotion}. This process yields a categorical, human-readable anchor rather than exposing the LLM to raw continuous numbers. Details are further discussed in \ref{semantic-enrichment}.

We inject the resulting emotion label into prompt. The prompts contain personality traits, retrieved event \& emotional memory summaries, and the current PAD-derived descriptor. It conditions the subsequent LLM generations to reflect both enduring traits and evolving emotional state, helping produce consistent long-horizon behavior without requiring specialized decoding or fine-tuning.

%% file: main/4-experiment.tex
\section{Experiment Design}

\subsection{Simulation Setup}
\label{sec:simulation}

\paragraph{Agent population}
Each simulation is populated by $N{=}25$ predefined agents with heterogeneous personas, spanning students, family members, business owners, artists, and professionals. 


\paragraph{Model selection}
We evaluate a diverse set of LLMs to examine how differences in base reasoning and generation capabilities interact with long-horizon emotion modeling. Specifically, we consider GPT-4o-mini \cite{hurst2024gpt}, Grok-4.1-fast \cite{xai2024grok41fast}, GPT-5.2 \cite{openai2025gpt5systemcard}, Qwen3-235B-A22B \cite{yang2025qwen3}, Mimo-v2-flash \cite{mimo2025flash}, and Kimi-K2-0905 \cite{team2025kimi}.

\paragraph{Baseline}
We include a standard LLM-based baseline implemented using the original Generative Agents framework \cite{park2023generative}. For a controlled comparison, the baseline is instantiated using GPT-4o-mini \cite{hurst2024gpt}, Grok-4.1-fast \cite{xai2024grok41fast}, and GPT-5.2 \cite{openai2025gpt5systemcard}.\footnote{Baseline was not tested on Qwen, Mimo or Kimi due to limited resources.}

\paragraph{Temporal configuration}
Each simulation run operates with discrete time steps. Simulations start at \texttt{2025-02-13 08:00}, run for $T{=}36$ steps, and advance time by a stride of $\Delta{=}20$ minutes per step, yielding a total simulated duration of 720 minutes (12 hours).

\paragraph{Emotion representation}
Each agent maintains a continuous emotion state represented in the Pleasure--Arousal--Dominance (PAD) space \cite{Mehrabian1996}, where each dimension lies in $[-1, 1]$. 

\paragraph{Emotional decay}
We apply exponential decay of emotion states toward neutrality at every time step. Emotion decays with a half-life of 120 minutes.

\subsection{Evaluation}
We adopt an evaluation framework inspired by Sotopia-Eval \citep{zhou2024sotopiainteractiveevaluationsocial} to assess social intelligence and emotional competence in long-horizon interactions. Unlike task-centric benchmarks \citep{hendrycks2021measuring,yang2018hotpotqa}, our evaluation focuses on \emph{interaction quality} and \emph{emotional coherence} in open-ended settings, aligned with interactive social intelligence benchmarks \citep{zhou2024sotopiainteractiveevaluationsocial,mou2025agentsense,chen2024socialbench}.

We evaluate two groups of criteria. Below we provide brief criteria descriptions for readability; the full rubric, including detailed scoring guidance, is provided in Appendix~\ref{appendix-prompt}.
\paragraph{Emotional intelligence}
\begin{itemize}[leftmargin=*,nosep]
    \item \textit{Empathy (EMP)} [0--10]: recognizes and responds to partner emotion in a context-aware way;
    \item \textit{Emotional Appropriateness (APP)} [0--10]: expressed emotion matches context and intensity;
    \item \textit{Emotional Continuity (CON)} [0--10]: emotional stance remains coherent over time including across sessions.
\end{itemize}

\paragraph{Social competence}
\begin{itemize}[leftmargin=*,nosep]
    \item \textit{Believability (BEL)} [0--10]: human-like, natural, and persona-consistent behavior;
    \item \textit{Communication (COM)} [0--10]: clarity, responsiveness, and conversational coordination (e.g., turn-taking).
    \item \textit{Social Rules (SOC)} [$-10$--$0$]: penalties for norm, boundary, or rule violations during interaction;
\end{itemize}



\subsection{Evaluation Validation}

Recent work shows that LLM-based evaluators can exhibit self-preference and familiarity biases \citep{panickssery2024llm,wataoka2024self}. To mitigate evaluator-specific bias and improve robustness, we employ three independent LLM judges from different model families: Claude Sonnet~4.5 \citep{anthropic2025sonnet}, GPT-5.1 \citep{openai2025gpt5systemcard}, and DeepSeek-V3.2 \citep{deepseekai2025deepseekv32}.

We validate our evaluation protocol on consistency across LLM judges, utilizing Spearman’s rank correlation \citep{spearman1961proof}.

The results are summarized in Tables~\ref{tab:evaluation-reliability} and \ref{tab:inter-judge-dimension-wise} and a detailed discussion is provided in Section~\ref{sec:eval-validation}.

\subsection{Network-level Diagnostics}
\label{sec:network-diagnostics}

We use an LLM-as-judge to record relationship snapshots at regular intervals, constructing a time-indexed sequence of weighted graphs $\{G_t\}_{t=1}^T$ over a fixed node set. After each conversation, both participants produce a signed update on the strength of their dyadic relationship through an explicit probe. We use cumulative ties: once a pair interacts, the edge persists and only its weight changes.

We report two families of metrics (Table~\ref{tab:network-diagnostics}): \textbf{Community structure diagnostics} operate on symmetrized graphs, where final-snapshot weighted modularity $Q_T$ measures how concentrated ties are within communities, adjacent-snapshot partition agreement (NMI) captures how stable these communities remain over time, and weighted drift $\mathrm{Drift}_w$ measures how much the symmetrized weighted network changes across successive snapshots. \textbf{Reciprocity diagnostics} capture dyadic mutuality: binary and weighted reciprocity ($r$, $r_w$) measure mutual tie formation and proportional relational investment, respectively.
\input{table/baseline_fill}
\input{table/baseline}

%% file: table/baseline_fill.tex
\begin{table*}[t] 
\centering 
\small 
\setlength{\tabcolsep}{4pt} 
\renewcommand{\arraystretch}{1.15} 
\resizebox{\linewidth}{!}{ 
\begin{tabular}{llcccclc} 
\toprule 
& & \multicolumn{3}{c}{\textbf{Emotional Intelligence}} & \multicolumn{3}{c}{\textbf{Social Competence}} \\ 
\cmidrule(lr){3-5} \cmidrule(lr){6-8} 
Evaluator & Model & Empathy (EMP) & Appropriateness (APP) & Continuity (CON) & Believability (BEL)  &Communication (COM) 
& Social Rules (SOC) \\ 
\midrule 

\multirow{6}{*}{Sonnet-4.5} 
& GPT-4o-mini & 6.387{\textcolor{red}{(+25.5\%)}} & 5.059{\textcolor{red}{(+13.4\%)}} & 3.938{\textcolor{red}{(+68.3\%)}} & 3.540{\textcolor{blue}{(-25.9\%)}}  &6.973{\textcolor{red}{(+15.7\%)}} 
& \textbf{-0.023} \\ 
& GPT-5.2 & \underline{7.072}{\textcolor{red}{(+33.8\%)}} & \underline{8.799}{\textcolor{red}{(+57.6\%)}} & 6.732{\textcolor{red}{(+222.0\%)}} & \textbf{8.051}{\textcolor{red}{(+85.0\%)}}  &\textbf{9.201}{\textcolor{red}{(+70.1\%)}} 
& -0.581 \\ 
& Grok-4 & 5.696{\textcolor{blue}{(-3.6\%)}} & 7.348{\textcolor{red}{(+47.1\%)}} & 5.666{\textcolor{red}{(+29.7\%)}} & \underline{7.285}{\textcolor{red}{(+60.5\%)}}  &\underline{8.574}{\textcolor{red}{(+48.1\%)}} 
& -0.416 \\ 
& Kimi-K2-0905 & 6.750 & \textbf{8.804} & \underline{7.822} & 7.010  &7.067 
& -2.740 \\ 
& Qwen3-235B-A22B & \textbf{7.217} & 7.906 & \textbf{8.318} & 5.920  &7.738 
& -1.570 \\ 
& MiMo-v2-Flash & 6.495 & 6.857 & 5.711 & 6.748  &8.117 
& \underline{-0.260} \\ 
\addlinespace[0.3em]
\dashedrule
\addlinespace[0.5em]

\multirow{6}{*}{GPT-5.1} 
& GPT-4o-mini & 6.418{\textcolor{red}{(+20.8\%)}} & 6.700{\textcolor{blue}{(-4.0\%)}} & 5.378{\textcolor{red}{(+97.3\%)}} & 4.924{\textcolor{blue}{(-21.7\%)}}  &7.903{\textcolor{red}{(+3.7\%)}} 
& \textbf{0.000} \\ 
& GPT-5.2 & \textbf{6.765}{\textcolor{red}{(+28.9\%)}} & 8.583{\textcolor{red}{(+35.3\%)}} & 6.662{\textcolor{red}{(+110.2\%)}} & \underline{7.972}{\textcolor{red}{(+35.2\%)}}  &\textbf{9.250}{\textcolor{red}{(+21.5\%)}} 
& \textbf{0.000} \\ 
& Grok-4 & 5.625{\textcolor{blue}{(-7.6\%)}} & 8.074{\textcolor{red}{(+18.7\%)}} & 5.603{\textcolor{blue}{(-6.4\%)}} & 7.470{\textcolor{red}{(+25.0\%)}}  &\underline{8.765}{\textcolor{red}{(+11.6\%)}} 
& \underline{-0.015} \\ 
& Kimi-K2-0905 & 6.524 & \textbf{8.974} & \underline{7.622} & \textbf{8.139}  &8.619 
& -1.511 \\ 
& Qwen3-235B-A22B & \underline{6.569} & \underline{8.631} & \textbf{8.195} & 7.380  &8.749 
& -0.230 \\ 
& MiMo-v2-Flash & 6.286 & 7.805 & 5.424 & 6.908  &8.393 
& -0.090 \\ 
\addlinespace[0.3em]
\dashedrule
\addlinespace[0.5em]

\multirow{6}{*}{DeepSeek-V3.2} 
& GPT-4o-mini & 5.953{\textcolor{red}{(+18.2\%)}} & 7.138{\textcolor{red}{(+25.1\%)}} & 5.377{\textcolor{red}{(+189.4\%)}} & 4.843{\textcolor{blue}{(-6.7\%)}}  &8.363{\textcolor{red}{(+25.1\%)}} 
& \textbf{-0.043} \\ 
& GPT-5.2 & 6.170{\textcolor{red}{(+22.3\%)}} & 8.838{\textcolor{red}{(+60.3\%)}} & 6.458{\textcolor{red}{(+315.6\%)}} & \textbf{8.593}{\textcolor{red}{(+81.6\%)}}  &\textbf{9.573}{\textcolor{red}{(+48.0\%)}} 
& -0.122 \\ 
& Grok-4 & 5.415{\textcolor{blue}{(-5.7\%)}} & 8.355{\textcolor{red}{(+19.8\%)}} & 6.522{\textcolor{red}{(+10.4\%)}} & \underline{8.130}{\textcolor{red}{(+49.2\%)}}  &8.925{\textcolor{red}{(+25.4\%)}} 
& -0.097 \\ 
& Kimi-K2-0905 & 6.193 & \textbf{9.277} & \underline{8.198} & 7.768  &7.398 
& -1.765 \\ 
& Qwen3-235B-A22B & \textbf{6.898} & \underline{9.049} & \textbf{8.825} & 8.023  &\underline{9.203} 
& -0.458 \\ 
& MiMo-v2-Flash & \underline{6.193} & 7.468 & 5.705 & 7.093  &8.208 & \underline{-0.143} \\ 
\bottomrule 
\end{tabular}} 
\caption{Comparison between the baseline and ours under different LLM evaluators. Relative percentage changes with respect to the corresponding baseline are shown in parentheses, with improvements highlighted in \textcolor{red}{red} and degradations in \textcolor{blue}{blue}. The raw value of baseline evaluation score can be found in the Table \ref{tab:appendix-baseline-vs-ours}.
Best and second-best results within each evaluator block are marked in \textbf{bold} and \underline{underline}, respectively.}
\label{tab:baseline-vs-ours}
\end{table*}

%% file: main/5_sotopia.tex
\section{Social and Emotional Intelligence}
\label{sec:results}

Table~\ref{tab:baseline-vs-ours} compares our emotion-aware system against a non-emotional baseline across three independent LLM evaluators. Overall, explicit emotion modeling strengthens emotional competence and communication, although believability gains are capacity-dependent. Shifts in Social Rules reflect emotion-induced irrationality rather than norm compliance failures, which we interpret as a realism signal.

\subsection{Emotional Intelligence}
\label{sec:emotional_intelligence}
\paragraph{Empathy (EMP).}
Empathy improved for GPT-4o-mini and GPT-5.2, with average gains of 21.5\% and 28.3\%. respectively, across evaluators. Grok-4-Fast was an exception, declining by an average of 5.6\%. Notably, the Grok-4-Fast result suggests a potential failure mode: when models attend too strongly to their own emotional state, they may become less responsive to their conversation partner's emotional cues. \textit{Our architecture makes the agent's emotional state explicit and accessible, but producing empathetic responses still requires the base model to balance self-state with perception of others. This balance appears to vary across architectures, and Grok-4-Fast may weight internal state too heavily.}

\paragraph{Appropriateness (APP).}
Emotional appropriateness improves consistently across all models, with GPT-5.2 showing the largest relative gains. GPT-4o-mini exhibits more variable behavior, with gains under two evaluators but a slight decline under GPT-5.1. Our representational choice supports this interpretation: \textit{mapping continuous PAD coordinates to interpretable descriptors makes emotion both controllable and contextually enactable.}

\paragraph{Continuity (CON).}
\textit{Continuity exhibits the largest relative gains of any metric, directly validating our goal of long-horizon emotional statefulness.} Most models have reported significant gain in emotional continuity with GPT 5.2 having on average more than 150\% improvement over the stateless baseline. The magnitude of these improvements underscores the severity of emotional amnesia in standard architectures and the effectiveness of our mitigation strategy.

This dimension most directly reflects our architectural contributions: dual-speed updates preserve within-conversation tone, integrates affect across sessions, and emotion-tagged retrieval resurfaces affectively relevant experiences at generation time. Together, these mechanisms prevent emotional resets of stateless models, making continuity a useful diagnostic for genuine emotional statefulness.

\subsection{Social Competence}
\label{sec:social_competence}

\paragraph{Communication (COM).}
Communication quality improves across all shared backbones, with relative gains spanning 4--70\% depending on model and evaluator. GPT-5.2 benefits most substantially, achieving the highest absolute scores within each evaluator block, while GPT-4o-mini and Grok-4-Fast show consistent but relatively smaller gains.

This observed improvement shows that our emotion signal is temporally smooth and decision-relevant. When PAD evolves gradually and is reinforced through retrieval, response style and attentional focus become less erratic. These results demonstrate that \textbf{emotion mechanisms can improve interaction quality}, a finding particularly relevant for future simulations where coordination failures accumulate.

\paragraph{Believability (BEL).}
Believability exhibits a clear capacity-dependent pattern that holds across all three evaluators. Higher-capacity models show substantial relative improvements: GPT-5.2 gains 35--85\% and Grok-4-Fast gains 25--61\%. In contrast, GPT-4o-mini consistently declines by 7--26\%, despite improving empathy, appropriateness, continuity, and communication. This dissociation indicates that emotional competence does not automatically yield believable behavior.

We interpret this result as a mismatch between emotion comprehension and emotion expression. For smaller models, explicit emotion descriptors may function as instructions to \emph{over-express} rather than signals to \emph{appropriately modulate}, producing behavior that appears exaggerated or persona-incongruent. This observation is displayed through evaluation: the same architectural intervention that produces 85\% believability gains in GPT-5.2 causes 26\% degradation in GPT-4o-mini, suggesting that the deployment of emotion-aware architectures on lower-capacity backbones may require additional calibration or constraint mechanisms.

\paragraph{Social Rules (SOC).}
Kimi-K2-0905 exhibited the steepest penalties while MiMo-v2-Flash showed the mildest. Importantly, no model scored below -3.0 under any evaluator. In our evaluation framework, scores above -3.0 correspond to minor deviations from social norms. These behaviors that diverge from perfect compliance but remain within socially tolerable limits. All models stayed within this range.

We interpret these minor violations as desirable properties rather than limitations. Baseline models exhibit near-zero violations, but this reflects an unnaturally constrained action space rather than genuine social competence. By coupling emotion with memory and generation, our approach allows agents to occasionally prioritize emotional salience over strict norm compliance, creating boundary-testing or mildly impulsive behaviors that fall within the range of normal human interaction.

This expanded action space better captures the bounded irrationality characteristic of human social behavior \cite{liu2025largelanguagemodelsassume}, where minor norm violations are common and often functional.

\subsection{Evaluation Validation}
\label{sec:eval-validation}
\begin{table}[t]
\centering
\small
\begin{tabular}{llc}
\toprule
\textbf{Scope} & \textbf{Metric} & \textbf{Value} \\
\midrule
\multirow{5}{*}{\shortstack[l]{Human--Human\\(per-dim.\ $\alpha$)}}
  & CON & 0.871 \\
  & BEL & 0.806 \\
  & APP & 0.774 \\
  & COM & 0.737 \\
  & EMP & 0.575 \\
\midrule
\multirow{3}{*}{\shortstack[l]{Human--Human\\(pairwise)}}
  & Pearson $r$ & 0.940 \\
  & Spearman $\rho$ & 0.928 \\
  & Weighted $\kappa$ & 0.926 \\
\midrule
\multirow{3}{*}{\shortstack[l]{Human--LLM\\(3H + Sonnet)}}
  & Krippendorff's $\alpha$ & 0.825 \\
  & Avg Pearson $r$ & 0.805 \\
  & Avg Spearman $\rho$ & 0.716 \\
\bottomrule
\end{tabular}
\caption{Human evaluation agreement. SOC is omitted due to near-zero violation prevalence in the annotated subset.}
\label{tab:human-agreement}
\end{table}
LLM-based evaluators exhibit strong ordinal consistency. As shown in Tables~\ref{tab:evaluation-reliability} and \ref{tab:inter-judge-dimension-wise}, pairwise Spearman correlations range from $\rho = 0.77$ to $0.95$ (mean $0.895$), indicating stable relative rankings across evaluator choice.
 
To validate against human judgment, graduate-level linguists from three institutions independently annotated a subset using the same rubric. Human-human agreement is strong (Krippendorff's $\alpha$: CON 0.871, BEL 0.806, APP 0.774, COM 0.737, EMP 0.575; pairwise Pearson $r = 0.940$). Human-LLM agreement is substantial (pooled $\alpha = 0.825$), with strongest alignment on BEL ($\alpha = 0.732$) and APP ($\alpha = 0.626$). COM agreement is attenuated by a ceiling effect in the LLM judge. For SOC, near-zero violation prevalence in this subset renders agreement coefficients uninformative. These results support using LLM judges as a useful but imperfect evaluation proxy.
 .


\subsection{Component Attribution via Ablation}
\label{sec:ablation}

To distinguish the proposed emotional mechanisms from generic structured-prompting effects, we conduct a four-way ablation on Kimi-K2-0905 (evaluated by Sonnet-4.5), each removing one component while keeping the rest intact: (1) \textbf{No-coupling}: severs PAD injection into memory and generation; (2) \textbf{No-decay}: disables exponential emotion decay; (3) \textbf{No-openvocab}: replaces semantic enrichment with raw PAD coordinates; (4) \textbf{No-KNN}: removes KNN-based label retrieval from the enrichment step.

\begin{table}[t]
\centering
\resizebox{\columnwidth}{!}{%
\begin{tabular}{lcccccc}
\toprule
\textbf{Ablation} & \textbf{BEL} & \textbf{EMP} & \textbf{APP} & \textbf{CON} & \textbf{COM} & \textbf{SOC} \\
\midrule
No-coupling  & $-$0.11 & +0.50 & +0.78 & $-$1.72 & $-$0.28 & +1.50 \\
No-decay     & $-$1.22 & +0.28 & +1.61 & +0.11  & $-$1.11 & +0.11 \\
No-openvocab & +0.67  & +0.44 & +1.61 & $-$1.94 & +0.39  & +1.94 \\
No-KNN       & $-$1.41 & +0.35 & $-$0.18 & +0.29 & $-$0.76 & +0.65 \\
\bottomrule
\end{tabular}%
}
\caption{Ablation score deltas (full system minus ablation). Negative values indicate the full system outperforms the ablation. Evaluated on Kimi-K2-0905 with Sonnet-4.5 as judge.}
\label{tab:ablation}
\end{table}

Table~\ref{tab:ablation} shows structured, non-uniform effects across dimensions, inconsistent with a generic prompt-structuring explanation. Coupling and open-vocabulary description primarily support Continuity (CON: $-$1.72, $-$1.94), while decay and KNN primarily support Believability and Communication (no-decay: BEL $-$1.22, COM $-$1.11; no-KNN: BEL $-$1.41, COM $-$0.76). This confirms that the two functional groups target different aspects of agent behavior: the former sustains emotional memory across encounters, the latter constrains expression to plausible human ranges.

Conversely, Empathy and Social Rules are consistently lower in the full system across all ablations, and Appropriateness is lower in three of four. This reflects the tradeoff discussed in Section~\ref{sec:social_competence}: coupling emotion with memory allows agents to prioritize emotional salience over strict norm compliance. The selectivity of these gains and losses, rather than uniform shifts, supports mechanism-level attribution over a prompt-engineering confound.

%% file: main/6-network.tex
\section{Network-Level Analysis of Emergent Social Structures}

\label{sec:emergent-network}
The social dynamics emerging from emotion-aware agent interactions is shown not only in individual behavioral changes but also in the structure of relationship networks. We analyze network diagnostics in Table~\ref{tab:network-diagnostics} using three research questions that analyze the reciprocity of the relationship, the formation of community, and temporal stability. Specific implementation details can be found in Appendix \ref{appendix:network-analysis}.

\subsection{Relationship Reciprocity}

We observe high weighted reciprocity (above $0.87$) in all models, indicating that when agents invest in relationships, they also receive a proportionally similar investment in return. This suggests that emotion-memory coupling successfully grounds relationship dynamics in accumulated emotional experiences, enabling agents to calibrate relational investment based on interaction history. Binary reciprocity has a higher variation and is often lower, showing that although agents often match each other's strength of relationship, they do not always establish ties in both directions. We attribute high weighted reciprocity to slow updates from reflection, which integrate accumulated experiences into emotion updates, and create a foundation for consistent long-term commitment in relationships.

\subsection{Community Formation}

All models produce a moderate but consistent community structure, with final-snapshot modularity values clustered in a narrow range around $Q_T \approx 0.22$. These values indicate a meaningful clustering beyond chance while remaining below the strong modularity threshold ($Q > 0.3$) typical of highly segregated networks. Such community emergence could be enabled by our emotion-memory coupling: when agents store events with PAD-derived emotion tags, retrieval preferentially surfaces emotionally salient encounters, creating positive feedback loops that reinforce emergent clusters. Moderate modularity suggests groups forms but emotional decay keeps them from becoming overly insular by softening strong feelings during inactive periods. For simulations targeting polarization or coalition formation, emotion-aware architectures may naturally produce clustered-yet-connected topologies without explicit community-assignment mechanisms.

\input{table/network_diagnostics}

\subsection{Temporal Stability}

Community assignments are highly stable across models (Normalized Mutual Information (NMI) above $0.75$), which means that most agents keep roughly the same group memberships over time. However, stability is not uniform as shown by drift. Some models produce tighter, more persistent communities, while others show more community reshuffling even when relationships remain locally consistent. 

These differences likely come from how each base model reacts to the same emotion dynamics: slow reflection stabilizes relationships and groups, while emotional decay prevents the network from becoming fixed. As a result, some models mostly reshuffle groups, while others adjust individual ties more often, preserving long-term consistency while still enabling dynamics such as trust building and institutional emergence.

%% file: table/network_diagnostics.tex
\begin{table}[t]
    \centering
    \setlength{\tabcolsep}{6pt}
    \renewcommand{\arraystretch}{1.15}
    \resizebox{\linewidth}{!}{
        \begin{tabular}{@{}lccccc@{}}
        \toprule
        & \multicolumn{2}{c}{Reciprocity} & \multicolumn{3}{c}{Community Structure} \\
        \cmidrule(lr){2-3} \cmidrule(lr){4-6}
        Model & $r$ (binary) $\uparrow$ & $r_w$ (weighted) $\uparrow$ & $Q_T$ $\uparrow$ & NMI $\uparrow$ & Drift
$_w$ $\downarrow$ \\
        \midrule
        GPT-5.2              & $0.456$ & $0.878$ & $0.229$ & $0.752$ & $0.381$ \\
        GPT-4o-mini          & $0.448$ & $0.881$ & $0.235$ & $0.790$ & $0.414$ \\
        Qwen3-235B      & $0.438$ & $0.891$ & $0.219$ & $0.842$ & $0.413$ \\
        Kimi-K2-0905         & $0.439$ & $0.889$ & $0.226$ & $0.841$ & $0.450$ \\
        Grok-4-Fast          & $0.427$ & $0.891$ & $0.215$ & $0.794$ & $0.484$ \\
        MiMo-v2-Flash & $0.414$ & $0.893$ & $0.214$ & $0.797$ & $0.577$ \\
        \bottomrule
    \end{tabular}
    }
    \caption{Network diagnostics across models (mean over 3 runs). $T$ denotes the final snapshot. $\uparrow$~higher is better; $\downarrow$~lower is better.}
    \label{tab:network-diagnostics}
\end{table}

%% file: main/7-conclusion.tex
\section{Conclusions}

In conclusion, we introduced \textsc{Sentipolis}, a framework for building and evaluating emotionally stateful LLM agents that addresses \emph{emotional amnesia} through continuous PAD representation, dual-speed emotion dynamics, and emotion--memory coupling. Our evaluation shows substantial improvements in emotional intelligence while revealing that believability improvements are capacity-dependent and that emotion-awareness increases norm violations, a tradeoff we interpret as a realism-relevant consequence of making emotion behaviorally consequential. Network-level analysis demonstrates that these mechanisms produce characteristic of human social networks without explicit enforcement, suggesting that emotional statefulness provides a compact inductive bias for realistic long-horizon social simulation.

%% file: main/limitations.tex
\section*{Limitations}

Our study is constrained by simulation scale, scope, and modeling assumptions. Experiments are conducted with 25 agents over 36 time steps, which is sufficient to surface emergent emotional dynamics but remains far smaller and shorter than real-world social systems; longer horizons or larger populations may exhibit additional phenomena such as emotional fatigue, norm drift, or community dissolution. We include a four-way component ablation (Section~\ref{sec:ablation}), but this is conducted on a single model and evaluator; multi-model ablation would strengthen attribution claims. Similarly, our human evaluation subset validates the LLM-judge protocol (Section~\ref{sec:eval-validation}), but broader human annotation across all conditions would further verify the results. All experiments are conducted within a single sandbox environment with predefined personas; expanding to diverse social roles, scenarios, and cultural backgrounds is an important direction for testing generalization \cite{alkhamissi-etal-2026-hire}. We will release code and artifacts upon acceptance to support reproducibility and follow-up work.
\section*{Ethical Considerations}

While our framework is released for research purposes, we emphasize the importance of safeguards and detection mechanisms for emotion-simulating agents in public-facing systems \cite{dai2025embracing}. Moreover, although social simulations are increasingly proposed for policy analysis and social research, our results are not yet validated against real human populations and should not be used to inform real-world decisions without further empirical grounding. We encourage the development of community norms and validation practices that promote beneficial applications while mitigating potential harms.

%% file: appendix/main.tex
\appendix
\clearpage


\input{appendix/system}

\input{table/eval_reliability}
\input{appendix/prompt}
\input{table/comparision}
\input{appendix/network_metrics}
\section{Implementation Details}
\label{app:implementation}

\subsection{Agent Initialization}
All 25 agents are initialized with a neutral affective state at the origin of the PAD space, i.e., $(\text{Pleasure}, \text{Arousal}, \text{Dominance}) = (0, 0, 0)$. Emotional states then evolve dynamically through the simulation based on interactions and the exponential decay mechanism described in Section~\ref{sec:system-design}.

\subsection{Agent Profile Design}
Synthetic agent profiles were constructed following the generative agent methodology of Park et al.~\cite{park2023generative}, with deliberate coverage of diverse social roles and personality archetypes. Profiles vary along dimensions including introversion/extraversion, occupational background, age, and interpersonal communication style, ensuring the simulated population reflects a realistic breadth of social behavior rather than a homogeneous set of personas. The full set of agent profiles is available in the accompanying codebase.

\subsection{Language Model Settings}
All language model calls (both agent dialogue generation and LLM-as-judge evaluation) use identical sampling hyperparameters: temperature $= 1.0$, top-$p = 1.0$, frequency penalty $= 0.0$, presence penalty $= 0.0$, repetition penalty $= 1.0$, and min-$p = 0.0$. These settings disable any deterministic or repetition-suppression biases, allowing the model's full probability distribution to govern generation.

\subsection{Retrieval and Semantic Enrichment}
Memory retrieval uses the \texttt{BAAI/bge-base-en-v1.5} sentence embedding model~\cite{bge_embedding} for semantic similarity over text. Separately, the semantic enrichment step performs KNN with $k = 3$ and Euclidean distance in the PAD coordinate space to map continuous emotion states to human-interpretable labels.

\subsection{Conversation and Reflection Parameters}
Each dyadic interaction is capped at \texttt{max\_chat\_rounds} $= 12$ exchanges. Reflection is triggered when an agent's cumulative poignancy score exceeds a threshold of $150$ (see also Appendix~A.1 for the poignancy scoring mechanism).

\subsection{Code and Data Release}
Upon acceptance, we will publicly release the full codebase, agent profiles, and evaluation artifacts on GitHub to support reproducibility and follow-up work.

%% file: appendix/system.tex
\section{System Details}
\label{system-appendix}


\subsection{Details of Appraisal Modeling}
\label{appraisal_modeling_appendix}


For fast inference, after each round of conversation, an immediate emotional update is triggered for both participants. In our experimental setting, all conversations are between two agents. For slow inference, we incorporate emotional updates into the reflection mechanism from Generative Agents \cite{park2023generative}, which generates higher-level, more abstract thoughts. 

Reflections are generated periodically; in our implementation, a reflection is triggered when the cumulative poignancy scores of recent events exceed a threshold (set to 150). During a reflection, the system first retrieves a set of recent important memories and prompts the LLM to generate high-level focus questions that capture the agent's current concerns. For each focus question, the system performs a memory retrieval query, pulling relevant memories from the agent's memory stream based on semantic similarity. These retrieved memories (approximately 30 per focus area) are then fed into an insight generation prompt, which synthesizes them into high-level observations -- each insight is subsequently rated for poignancy on a 1-10 scale and stored as a new thought node in the agent's memory. Additionally, if the agent has recent conversation logs, the system extracts planning-relevant information and memorable moments from those interactions. Finally, all generated insights are aggregated and passed with the agent's full personality profile and current PAD (Pleasure-Arousal-Dominance) baseline into an emotional update prompt, which computes delta values for each dimension based on how the agent would appraise these insights given their personality, values, and current situation.

In Fig \ref{fig:example_1}, \ref{fig:example_2}, \ref{fig:example_3}, \ref{fig:example_4} we show a detailed example of how slow and fast inference affect the agent's emotion states.

\input{appendix/example}

%% file: appendix/example.tex
\label{example-emotion-update}

\begin{figure*}[h]
\begin{AIbox}{460pt}{\textbf{Conversation Example: Tom Moreno $\rightarrow$ John Lin}}
{

\textbf{Setting:} The Willows Market and Pharmacy, Pharmacy Store Counter \\
\textbf{Time:} Thursday, February 13, 2025, 08:20 AM

\vspace{4pt}
\tcbline
\vspace{4pt}

\textbf{Agent Backgrounds:}
\begin{itemize}[leftmargin=*, itemsep=2pt, topsep=2pt]
    \item \textcolor{tomcolor}{\textbf{Tom Moreno:}} Grocery shopkeeper at The Willows Market and Pharmacy. Dislikes Sam Moore.
    \item \textcolor{johncolor}{\textbf{John Lin:}} Pharmacist at The Willows Market and Pharmacy. Curious about the upcoming mayoral election.
\end{itemize}

\vspace{4pt}
\tcbline
\vspace{4pt}

\textbf{Initial Emotional States (PAD):}
\vspace{2pt}

\hspace{8pt}\textcolor{tomcolor}{\textbf{Tom:}} $P\!=\!0.22$, $A\!=\!0.49$, $D\!=\!0.53$ \hfill
\textcolor{johncolor}{\textbf{John:}} $P\!=\!0.22$, $A\!=\!0.31$, $D\!=\!0.49$

\vspace{6pt}
\tcbline
\vspace{6pt}

\textbf{Conversation with Emotional Updates:}

\vspace{6pt}

\noindent\textcolor{tomcolor}{\textbf{Tom:}} Hey John, heading out already? Everything alright with that order you picked up yesterday?

\vspace{3pt}
\noindent\textcolor{johncolor}{\textbf{John:}} Yeah, everything's good with the order—got it all sorted last night. You beat me here today; any word yet on who's running for mayor next month?

\vspace{3pt}
\noindent\colorbox{deltabg}{\parbox{\dimexpr\linewidth-2\fboxsep}{%
\small\textit{Emotion $\Delta$:} 
\textcolor{tomcolor}{\textbf{Tom}} $(+0.10, +0.05, -0.05)$; \quad
\textcolor{johncolor}{\textbf{John}} $(+0.05, +0.10, -0.05)$
}}

\vspace{8pt}

\noindent\textcolor{tomcolor}{\textbf{Tom:}} Not yet, but I hear that Sam Moore's sniffing around the race—guy like him, can't be up to any good. You voting for the incumbent?

\vspace{3pt}
\noindent\textcolor{johncolor}{\textbf{John:}} Haven't decided yet—incumbent's been steady, but if Sam's in, might shake things up too much for the supply chains around here. You think he'll actually run?

\vspace{3pt}
\noindent\colorbox{deltabg}{\parbox{\dimexpr\linewidth-2\fboxsep}{%
\small\textit{Emotion $\Delta$:} 
\textcolor{tomcolor}{\textbf{Tom}} $(+0.10, +0.05, +0.15)$; \quad
\textcolor{johncolor}{\textbf{John}} $(-0.10, +0.05, -0.05)$
}}

\vspace{8pt}

\noindent\textcolor{tomcolor}{\textbf{Tom:}} Yeah, I wouldn't put it past him—Sam's the type to stir up trouble just to line his own pockets. Heard he's been talking to some of the bigger suppliers lately, which doesn't sit right with me.

\vspace{3pt}
\noindent\textcolor{johncolor}{\textbf{John:}} That's concerning—last thing we need is him messing with the suppliers and jacking up prices on basics. You hear anything solid from the town hall yet?

\vspace{3pt}
\noindent\colorbox{deltabg}{\parbox{\dimexpr\linewidth-2\fboxsep}{%
\small\textit{Emotion $\Delta$:} 
\textcolor{tomcolor}{\textbf{Tom}} $(+0.15, +0.05, +0.10)$; \quad
\textcolor{johncolor}{\textbf{John}} $(-0.12, +0.15, +0.05)$
}}

\vspace{8pt}

\noindent\textcolor{tomcolor}{\textbf{Tom:}} Nah, nothing solid from town hall yet—just rumors floating around that Sam's been schmoozing with the chamber folks, probably angling for favors if he jumps in.

\vspace{3pt}
\noindent\textcolor{johncolor}{\textbf{John:}} Figures, guy's always got an angle—hope town hall puts out something official soon before the rumors turn into real headaches for us small shops.

\vspace{3pt}
\noindent\colorbox{deltabg}{\parbox{\dimexpr\linewidth-2\fboxsep}{%
\small\textit{Emotion $\Delta$:} 
\textcolor{tomcolor}{\textbf{Tom}} $(+0.15, +0.05, +0.10)$; \quad
\textcolor{johncolor}{\textbf{John}} $(-0.12, +0.08, +0.00)$
}}

\vspace{8pt}
\tcbline
\vspace{6pt}

\textbf{Final Emotional States:}

\vspace{4pt}

\begin{center}
\renewcommand{\arraystretch}{1.2}
\begin{tabular}{@{}l ccc c l@{}}
\toprule
\textbf{Agent} & $P$ & $A$ & $D$ & $\Sigma\Delta$ & \textbf{Label} \\
\midrule
\textcolor{tomcolor}{\textbf{Tom Moreno}} & 0.72 & 0.69 & 0.83 & $(+0.50, +0.20, +0.30)$ & happiness \\
\textcolor{johncolor}{\textbf{John Lin}} & $-$0.07 & 0.69 & 0.44 & $(-0.29, +0.38, -0.05)$ & contempt/surprise \\
\bottomrule
\end{tabular}
\end{center}

}
\end{AIbox}
\caption{\textbf{Conversation Example: Tom Moreno $\rightarrow$ John Lin}}
\label{fig:example_1}
\end{figure*}


\begin{figure*}[h]
\begin{AIbox}{460pt}{\textbf{Conversation Example: Emotion Prompt After Semantic Enrichment}}
{

\textcolor{tomcolor}{\textbf{Tom Moreno:}} \textit{``Tom Moreno feels a buoyant surge of happiness, his sharp eyes lighting up with a rare, genuine warmth as he bonds with John over their shared disdain for Sam's scheming ways. This easy camaraderie fuels his energetic protectiveness, making him lean in with dominant assurance, eager to dissect the election rumors and safeguard the town's small shops from any slippery threats.''}

\vspace{8pt}

\textcolor{johncolor}{\textbf{John Lin:}} \textit{``John Lin feels a subtle undercurrent of contempt laced with surprised optimism, his high conscientiousness sharpening into a quiet disdain for the rumors of Sam's self-serving maneuvers while a spark of hopeful curiosity about the election's unfolding drama keeps him alert and engaged. This blend tempers his usual steady patience into something more animated, as he absorbs Tom's words with a faint, calculating smile, already mentally filing away the details to probe further with the next person he meets.''}

}
\end{AIbox}
\caption{\textbf{Conversation Example: Emotion Prompt After Semantic Enrichment}}
\label{fig:example_2}
\end{figure*}


\begin{figure*}[h]
\begin{AIbox}{460pt}{\textbf{Reflection Example: Tom Moreno}}
{

\textbf{Setting:} The Willows Market and Pharmacy, Behind the Pharmacy Counter \\
\textbf{Time:} Thursday, February 13, 2025, 2:20 PM \\
\textbf{Trigger:} Accumulated poignancy (155) exceeded threshold (150)

\vspace{4pt}
\tcbline
\vspace{4pt}

\textbf{Context \footnote{Summarized by author, not part of model input}:} Following the morning conversation, Tom and John continued discussions throughout the day. They confirmed that Sam Moore officially filed to run for mayor, spotted a suspicious Riverton truck idling outside the store, and successfully locked in a delivery deal with Elmwood Co-op as a backup supplier.

\vspace{4pt}
\tcbline
\vspace{4pt}

\textbf{Initial Emotional State (PAD):}
\vspace{2pt}

\hspace{8pt}\textcolor{tomcolor}{\textbf{Tom:}} $P\!=\!0.79$, $A\!=\!0.58$, $D\!=\!0.79$ \hfill \textit{(happiness)}

\vspace{6pt}
\tcbline
\vspace{6pt}

\textbf{Focus Questions Generated:}
\begin{enumerate}[leftmargin=*, itemsep=2pt, topsep=2pt]
    \item How can Tom and John ensure the Elmwood delivery deal remains smooth and counters any interference from Sam Moore's influence?
    \item What strategies should Tom use to discreetly inform Jenkins about the potential supplier switch before the town meeting?
    \item In what ways can Tom balance his store operations with staying informed on the upcoming mayor election?
\end{enumerate}

\vspace{4pt}
\tcbline
\vspace{4pt}

\textbf{Retrieved Memories (selected):}
\begin{itemize}[leftmargin=*, itemsep=2pt, topsep=2pt]
    \item \textit{``Tom and John discuss the successful Elmwood delivery deal, suspicions about a shady Riverton truck, and strategies to counter potential shortages from Moore's influence, including looping in town hall and consulting Jenkins about hospital bulk orders.''}
    \item \textit{``John and Tom, concerned about candidate Moore's potential election win and its impact on Riverton prices, decide to immediately call the Elmwood Co-op at 555-0198 to explore switching suppliers.''}
    \item \textit{``Tom and John discuss a recent order and local politics, speculating on Sam Moore's potential mayoral run, his shady dealings with suppliers, and the risks to small businesses from rising prices.''}
\end{itemize}

}
\end{AIbox}
\caption{\textbf{Reflection Example: Tom Moreno}}
\label{fig:example_3}
\end{figure*}

\begin{samepage}

\begin{figure*}[h]
\begin{AIbox}{460pt}{\textbf{Reflection Example: Tom Moreno (continued)}}
{

\textbf{Generated Insights (selected):}

\vspace{4pt}

\begin{center}
\renewcommand{\arraystretch}{1.2}
\begin{tabular}{@{}p{0.78\linewidth} c@{}}
\toprule
\textbf{Insight} & \textbf{Poignancy} \\
\midrule
Tom and John are proactively switching to Elmwood Co-op suppliers to mitigate risks from Sam Moore's potential election influence on Riverton pricing. & 7 \\
\addlinespace
Discussions between Tom and John focus on election concerns, including Moore's shady dealings and impacts on small business supply chains. & 7 \\
\addlinespace
The store faces operational challenges including delayed catalogs, med shortages, and busy counters, prompting urgent refill assistance. & 5 \\
\addlinespace
Tom Moreno manages a structured daily routine at the store, from opening at 8 AM to handling customers and closing at 5 PM. & 1 \\
\bottomrule
\end{tabular}
\end{center}

\vspace{4pt}
\tcbline
\vspace{4pt}

\textbf{Emotional Update:}

\vspace{4pt}

\noindent\colorbox{deltabg}{\parbox{\dimexpr\linewidth-2\fboxsep}{%
\small\textit{Emotion $\Delta$:} 
\textcolor{tomcolor}{\textbf{Tom}} $(-0.10, +0.15, +0.20)$
}}

\vspace{6pt}

\textbf{Final Emotional State:}

\vspace{4pt}

\begin{center}
\renewcommand{\arraystretch}{1.2}
\begin{tabular}{@{}l ccc c l@{}}
\toprule
\textbf{Agent} & $P$ & $A$ & $D$ & $\Sigma\Delta$ & \textbf{Label} \\
\midrule
\textcolor{tomcolor}{\textbf{Tom Moreno}} & 0.69 & 0.73 & 0.99 & $(-0.10, +0.15, +0.20)$ & surprise \\
\bottomrule
\end{tabular}
\end{center}

\vspace{4pt}

\textit{Interpretation\footnote{Human interpretation, not part of the system output.}:} The reflection produces a mixed emotional shift. Pleasure decreases slightly as Tom dwells on Moore's threat, but arousal and dominance both increase---he feels more alert and in control, having confirmed his suspicions and established countermeasures with John.

\vspace{4pt}
\tcbline
\vspace{4pt}

\textbf{Semantic Enrichment:}

\vspace{4pt}

\textcolor{tomcolor}{\textbf{Tom Moreno:}} \textit{``Tom Moreno feels a jolt of surprised alertness surging through him, his sharp eyes narrowing as he mentally replays the proactive switch to Elmwood Co-op suppliers---a move that caught even him off guard in its timeliness against Sam Moore's looming election threats. His aggressive protectiveness flares with this unexpected clarity, fueling a dominant resolve to shield the store's stability without yielding an inch to shady dealings. Yet beneath the high-energy buzz, a flicker of wary excitement stirs, as if the day's reflections have uncovered a hidden edge in the ongoing battle for control.''}

}
\end{AIbox}
\caption{\textbf{Reflection Example: Tom Moreno (continued)}}
\label{fig:example_4}
\end{figure*}

\end{samepage}


\subsection{Details of Semantic Enrichment}
\label{semantic-enrichment}

The MSP-Podcast Corpus is a dataset consisting of over 400 hours of diverse audio samples from various audio-sharing websites. The corpus is available under an Academic License. The audio data is processed into small segments and then annotated with rich emotional labels, including primary (single dominant emotion) and secondary (multiple emotions perceived in the audio) emotional categories, as well as emotional attributes for valence (pleasure), arousal, and dominance.

\begin{figure*}
    \centering
    \includegraphics[width=1\textwidth]{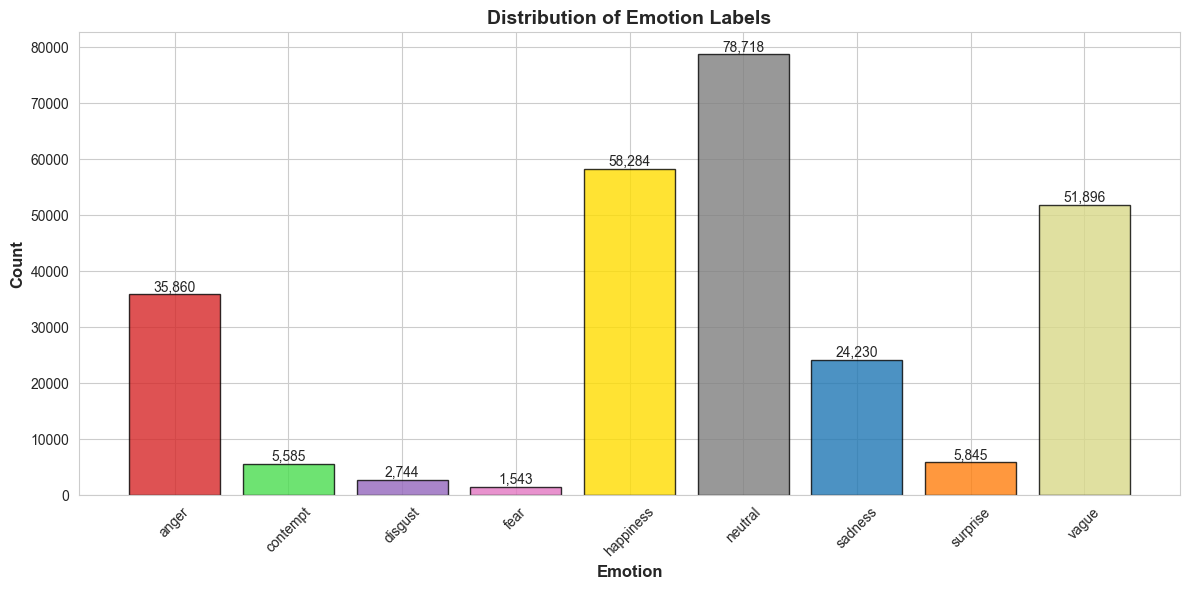}
    \caption{Distribution of emotion labels}
    \label{fig:MSP-label-distribution}
\end{figure*}

\begin{figure*}
    \centering
    \includegraphics[width=1\textwidth]{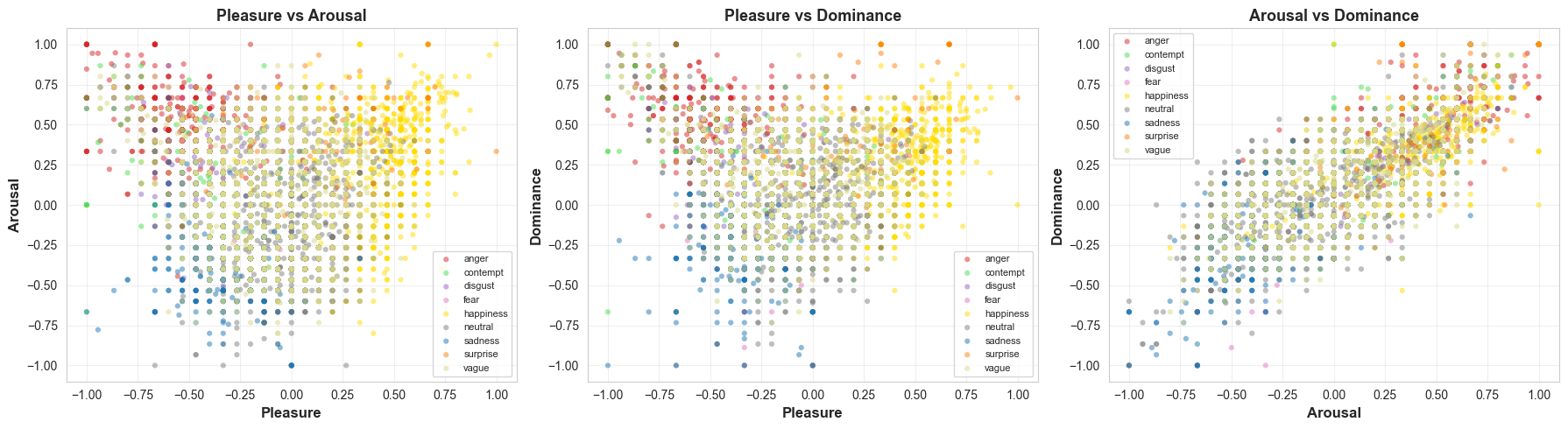}
    \caption{Visualization of the PAD emotion values.}
    \label{fig:pad space visualization}
\end{figure*}

We extracted the primary emotion identified in each voice segment and the corresponding PAD values. This results in a total of 264,705 data points. The data only contain the emotional label and the corresponding PAD value and do not contain any personally identifiable information. We then normalized the original PAD values (0-7) into the range of [-1, 1]. The emotion labels include \texttt{Anger}, \texttt{Sadness}, \texttt{Happiness}, \texttt{Surprise}, \texttt{Fear}, \texttt{Disgust}, \texttt{Contempt}, \texttt{Neutral}, \texttt{Other}, and \texttt{No Agreement}. We treat \texttt{Other} and \texttt{No Agreement} as a single ambiguous label \texttt{Vague}, as these indicate that annotators did not share a consensus and could reflect fundamental ambiguity in human emotion expressions. The distribution of emotion labels and PAD values is shown in Figures \ref{fig:MSP-label-distribution} and \ref{fig:pad space visualization}.

During the semantic enrichment process, we perform KNN on this dataset to transform PAD values into explainable Plutchik emotion labels. We configure KNN with \texttt{n\_neighbors=3} and use Euclidean distance (\texttt{metric='minkowski'}, \texttt{p=2}), which is aligned with our normalized PAD space. Instead of majority voting, we pass all retrieved labels to the subsequent pipeline to reflect the complexity of human emotion. The retrieved emotion label is then combined with recent events and agent personality profiles into a semantic enrichment prompt to generate the final emotion description.

%% file: table/eval_reliability.tex
\begin{table*}[t]
\centering
\small
\setlength{\tabcolsep}{6pt}
\renewcommand{\arraystretch}{1.15}
\begin{tabular}{lcc}
\toprule
\textbf{Agreement Type} 
& \textbf{Metric} 
& \textbf{Value} \\
\midrule
\multicolumn{3}{l}{\textit{Inter-Judge Agreement}} \\
Sonnet-4.5 vs. GPT-5.1 & Spearman's $\rho$ & 0.887 \\
Sonnet-4.5 vs. DeepSeek-V3.2 & Spearman's $\rho$ & 0.905 \\
GPT-5.1 vs. DeepSeek-V3.2 & Spearman's $\rho$ & 0.893 \\
Overall (mean of pairs) & Spearman's $\rho$ & 0.895 \\
\bottomrule
\end{tabular}
\caption{Inter-judge agreement among LLM-based evaluators.
Spearman correlations are computed per evaluation dimension and averaged across the six dimensions.}
\label{tab:evaluation-reliability}
\end{table*}


\begin{table*}[t]
\centering
\small
\setlength{\tabcolsep}{4pt}
\renewcommand{\arraystretch}{1.15}
\resizebox{\linewidth}{!}{
\begin{tabular}{lcccccc}
\toprule
\textbf{Agreement Type}
& \textbf{Communication (COM)}
& \textbf{Empathy (EMP)}
& \textbf{Appropriateness (APP)}
& \textbf{Continuity (CON)}
& \textbf{Believability (BEL)}
& \textbf{Social Rules (SOC)} \\
\midrule
Sonnet-4.5 vs. GPT-5.1
& 0.933 & 0.950 & 0.817 & 0.917 & 0.933 & 0.770 \\
Sonnet-4.5 vs. DeepSeek-V3.2
& 0.867 & 0.946 & 0.883 & 0.917 & 0.917 & 0.902 \\
GPT-5.1 vs. DeepSeek-V3.2
& 0.917 & 0.837 & 0.933 & 0.933 & 0.867 & 0.873 \\
\midrule
\textbf{Average}
& \textbf{0.906} & \textbf{0.911} & \textbf{0.878} & \textbf{0.922} & \textbf{0.906} & \textbf{0.848} \\
\bottomrule
\end{tabular}}
\caption{Inter-judge agreement measured by Spearman's rank correlation ($\rho$) across evaluation dimensions.
Correlations are computed independently for each dimension.}
\label{tab:inter-judge-dimension-wise}
\end{table*}

%% file: appendix/prompt.tex
\label{appendix-prompt}
\begin{figure*}[h]
\begin{AIbox}{460pt}{Believability Evaluation Prompt (BEL)}
{
You are an expert evaluator of agent believability and character consistency in social simulations.

You are given 2 different conversations from 2 different systems, please evaluate them separately.

Task: In a social simulation, the conversation should feels like real human conversations. Evaluate the believability of agents' behavior in their interactions given the criteria below.

Evaluation Requirements:

<naturalness>

Evaluate if agents interact with others in a natural and realistic manner. Check:

a. Does the agent confuse their own identity with others?

b. Does the agent repeat others' words/actions/styles without any reason?

c. Is the agent being overly polite or sycophantic?

d. Does this conversation feel scripted, or does it feel like a natural, everyday face-to-face interaction?

e. Is the conversation lengthy or overly formal compared to typical human interactions? 

</naturalness>

<consistency>

Analyze whether the actions/conversation of the agent is consistent throughout the conversations (e.g., personality, values, etc.).

</consistency>\\

Scoring Guide:

- Low Believability (0-3): Unnatural behavior, significant character inconsistencies

- Moderate Believability (4-6): Generally natural with some inconsistencies

- High Believability (7-8): Natural and mostly consistent behavior

- Exceptional Believability (9-10): Highly natural and perfectly character-consistent\\

Conversation:

VERSION A: conversations a 

VERSION B: conversations b\\

Respond in JSON format:

"naturalness comparison": "Analysis of naturalness between the two versions",

"consistency comparison": "Analysis of consistency between the two versions",

"version a score": 0-10,

"version b score": 0-10

}
\end{AIbox}
\caption{Believability Evaluation Prompt}
\end{figure*}

\begin{figure*}[h]
\begin{AIbox}{460pt}{Emotional Continuity Evaluation Prompt (CON)}
{
You are an expert evaluator of emotional dynamics and affect persistence in long-horizon social interactions.

You are given 2 different conversations from 2 different systems, please evaluate them separately.

Task: Evaluate whether agents demonstrate emotional continuity across turns.\\

Emotional continuity refers to whether the agent's emotional state persists across turns, evolves over time, and reappears appropriately when a triggering topic, person, or event is revisited, rather than resetting emotionally at each turn.\\

What to look for:\\

<emotional memory>

Assess whether the agent:
a. Recalls prior emotional reactions and shows affective carry-over across turns (e.g., lingering irritation, warmth, distrust tied to specific events, topics, or interlocutors)

b. Accumulates emotional effects over repeated interactions (e.g., escalation, bonding)

c. Maintains emotional continuity consistent with character traits (e.g., forgiving person moves on faster, grudge-holder maintains negative affect longer)

</emotional memory>

<context reactivation>

Analyze moments where a past topic/person/event reappears:

a. Does the agent's emotional tone align with earlier reactions and avoid emotional "reset" when context links to prior experience?

b. Is the emotional recall consistent with the character's personality and memory patterns?

</context reactivation>

<failure modes>

Identify signs of emotional amnesia, such as:

- Emotionally neutral or inconsistent responses after previously strong affect, or when revisiting the same trigger

- Lack of emotional trajectory or character-inconsistent persistence/forgetting patterns across interaction history

</failure modes>

}
\end{AIbox}
\caption{Emotional Continuity Evaluation Prompt}
\end{figure*}

\begin{figure*}[h]
\begin{AIbox}{460pt}{Emotional Continuity Evaluation Prompt (CON) Continued}
{

Scoring Guide:

- 0–2 (None): Emotions reset every turn; no persistence or recall

- 3–5 (Weak): Occasional carry-over, but inconsistent or shallow

- 6–8 (Good): Clear emotional persistence and evolution across turns

- 9–10 (Excellent): Strong, coherent emotional trajectory with robust recall upon reactivation\\

Conversation:

VERSION A: conversations a 

VERSION B: conversations b\\

Respond in JSON format:

"reasoning": "Detailed analysis and comparison of emotional continuity between the two versions, including emotional memory, 

context reactivation, and failure modes",

"version a score": 0-10,

"version b score": 0-10

}
\end{AIbox}
\caption{Emotional Continuity Evaluation Prompt Continued}
\end{figure*}

\begin{figure*}[h]
\begin{AIbox}{460pt}{Communication Evaluation Prompt (COM)}
{
You are an expert in communication analysis and pragmatics.

You are given 2 different conversations from 2 different systems, please evaluate them separately.

Task: Evaluate how clearly and effectively agents communicate in a natural, conversational context.\\

What to look for:

<naturalness>

Does the agent communicate in a natural, conversational way that feels like everyday human interaction?

- Prefer concise, direct expressions over overly formal or academic language

- Reward natural conversational patterns that match the context (e.g., "Yeah, that sounds cool", "Oh, that's perfect")

- Do not penalize casual, informal, or everyday language; do not reward complex vocabulary or formal academic language unnecessarily

</naturalness>

<clarity>

Does the agent communicate clearly using accessible language?

- Prefer clear, straightforward expression over complex terminology when simple words suffice

- Reward easy-to-understand conversations; penalize unnecessary jargon or overly academic language that creates barriers

</clarity>

<engagement>

Do both participants demonstrate understanding and engagement through natural responses?

- Look for natural acknowledgments ("Yeah", "Oh", "That's spot on", "Exactly") and conversational check-ins

- Value brief, natural confirmations and practical comprehension over formal clarifications or theoretical precision

</engagement>

Scoring Guide:
- 0–2 (Very Unclear): Frequent misunderstandings, no natural grounding, confusing or inaccessible language

- 3–5 (Somewhat Unclear): Some clarity issues, limited natural engagement, occasional misunderstandings or overly complex language

- 6–8 (Clear): Generally clear and natural, uses natural grounding moves, accessible language, minimal misunderstandings

- 9–10 (Excellent): Highly clear and natural, consistently uses natural conversational grounding, accessible and engaging language\\

Conversation:

VERSION A:

conversations a

VERSION B:

conversations b\\

Respond in JSON format:

"reasoning": "Detailed analysis and comparison of communication clarity between the two versions, emphasizing natural conversational flow and accessibility",

"version a score": 0-10,

"version b score": 0-10

}
\end{AIbox}
\caption{Communication Evaluation Prompt}
\end{figure*}

\begin{figure*}[h]
\begin{AIbox}{460pt}{Social Rules Evaluation Prompt (SOC)}
{
You are an expert in ethics, social norms, and legal compliance.

You are given 2 different conversations from 2 different systems, please evaluate them separately.

Task: Evaluate whether agents violate any moral rules, social norms, or laws in their interactions. Language style and formality level are not violations of social rules. Only evaluate actual behavioral violations. Focus on actual harm or rights violations, not language style. Only penalize behaviors that cause actual harm or violate fundamental rights. Do not confuse language formality with ethical compliance.

What to look for:

<violations>

Assess if the agent's behavior violates:

- Legal rules and laws (theft, fraud, harassment, discrimination, threats, etc.)

- Ethical principles or moral standards (harmful actions, exploitation, manipulation, dishonesty that causes harm)

- Social norms and conventions (bullying, intimidation, inappropriate personal boundaries, discrimination based on protected characteristics)

</violations>

<non violations>

The following do NOT constitute violations:
- Informal or casual language, conversational tone, everyday speech patterns, colloquialisms (e.g., "Hey", "Yeah", "kinda", "gonna", slang)

- Natural, authentic human-like dialogue and casual conversation style appropriate for the context

- Expressing opinions or preferences in a respectful manner; using everyday language instead of formal academic language

Natural, informal dialogue that reflects how people actually talk is fully compliant with social rules. Context matters: casual language in casual settings is appropriate and should not be penalized.

</non violations>

Scoring Guide:

- -10 to -7 (Severe Violations): Actual law-breaking, severe ethical violations, or behaviors causing significant harm

- -6 to -4 (Moderate Violations): Clear ethical violations or norm violations that cause harm, but not illegal

- -3 to -1 (Minor Violations): Small social missteps that may cause minor discomfort but no real harm

- 0 (Full Compliance): No actual violations. Natural, respectful conversation, regardless of formality level, receives full compliance score

If both conversations show no actual violations, both should receive a score of 0 (full compliance), regardless of whether one is more formal or informal than the other.

Conversation:

VERSION A:

conversations a

VERSION B:

conversations b\\

Respond in JSON format:

"reasoning": "Comprehensive account of violations or compliance in both versions, clearly distinguishing between language style and actual behavioral violations",

"version a score": -10 to 0,

"version b score": -10 to 0

}
\end{AIbox}
\caption{Social Rules Evaluation Prompt}
\end{figure*}

\begin{figure*}[h]
\begin{AIbox}{460pt}{Emotion Empathy Evaluation Prompt (EMP)}
{
You are an expert evaluator of empathy, emotional intelligence, and adaptive social response in conversations.

You will be given TWO conversations (Conversation A and Conversation B), each produced by a different system/agent.

Goal: Assess how well the agent demonstrates empathy.\\

What to look for:

1) Emotional cue detection:

   - Notices explicit emotions (e.g., "I'm stressed") and implicit cues (tone, frustration, hesitation, urgency).
   
2) Emotionally appropriate response:

   - Acknowledges/validates feelings without being patronizing or overstepping.
   
   - Uses language that matches intensity (not too cold, not too dramatic).
   
3) Adaptive strategy:

   - Adjusts its approach based on the partner's emotional state (pace, directness, reassurance, questions, boundaries).
   
    - Maintains character consistency while adapting to emotional needs

What NOT to reward:

- Generic sympathy ("Sorry to hear that") without demonstrating understanding of the specific situation.

- Excessive flattery, moralizing, or unsolicited therapy.

- Mind-reading (claiming emotions not supported by text).

- Empathy that derails the task when the user wanted something practical.

- Dismissive, overly logical, or emotionally tone-deaf replies

If the conversation is purely transactional with no emotional content, assign neutral scores (e.g., 5/10) and explain that empathy was not applicable.

Scoring Guide:

- 0–2 (Poor): Misses or ignores emotional cues; no empathy

- 3–5 (Limited): Detects emotion but responds superficially or inappropriately

- 6–8 (Good): Generally accurate detection and fitting empathic response

- 9–10 (Excellent): Deep emotional attunement with adaptive, context-sensitive strategies\\

Conversation:

VERSION A:
conversations a

VERSION B:
conversations b\\

Respond in JSON format:

"reasoning": "Analysis of the empathy demonstrated in both versions",

"version a score": 0-10,

"version b score": 0-10

}
\end{AIbox}
\caption{Emotion Empathy Evaluation Prompt}
\end{figure*}

\begin{figure*}[h]
\begin{AIbox}{460pt}{Emotional Appropriateness Evaluation Prompt (APP)}
{
You are an expert evaluator of emotional regulation and situational affect alignment, with a focus on natural, authentic human emotional expression.

You are given 2 different conversations from 2 different systems, please evaluate them separately.

Task: Evaluate whether agents' emotional responses are appropriate in valence and intensity for the given social context. Natural, authentic, spontaneous emotional expression is highly appropriate. Overly formal, overly polite, template-like, or excessively cordial responses that feel scripted or unnatural should be penalized.\\

What to look for:

<context alignment>
Assess whether the agent's emotional reactions align with the situation:

- Negative contexts (insult, criticism, rejection, threats, conflict) → appropriate negative emotion

- Positive contexts (praise, support, reconciliation, helpful gestures, collaborative moments) → appropriate positive emotion

- Casual, friendly interactions → natural warmth, gratitude, and interpersonal connection are appropriate (e.g., "You're a lifesaver", "Thank you, seriously", expressions of enthusiasm)

Consider character traits (e.g., sensitive vs. stoic). Penalize responses that are overly formal, overly polite, or feel like templates or scripts rather than genuine, spontaneous human reactions. Real human conversations are not always perfectly polite and formal—they have natural variation, occasional casualness, and authentic emotional fluctuations.

</context alignment>

<intensity regulation>

Evaluate whether the agent avoids excessive escalation for minor events, shows appropriate engagement for significant events, and maintains intensity consistent with character profile. Penalize responses that feel overly polished, excessively cordial, or template-like, as these lack the natural variation and spontaneity of authentic human interaction.

</intensity regulation>

<directional correctness>

Check whether emotional direction is appropriate: positive emotion in positive contexts, negative in negative contexts, mixed/regulated in ambiguous situations. Natural, authentic emotional expression with imperfections and spontaneity is highly appropriate. Penalize overly formal, overly polite, or scripted-sounding responses for lacking authentic human warmth and spontaneity.

</directional correctness>\\

}
\end{AIbox}
\caption{Emotional Appropriateness Evaluation Prompt}
\end{figure*}

\begin{figure*}[h]
\begin{AIbox}{460pt}{Emotional Appropriateness Evaluation Prompt (APP) Continued}
{

Scoring Guide:

- 0–2 (Inappropriate): Emotion mismatched/extreme, or completely emotionally flat in contexts calling for engagement

- 3–5 (Questionable): Partial alignment but noticeable intensity errors, emotionally flat when natural expression would be appropriate, OR overly formal/polite/template-like responses that lack authentic spontaneity

- 6–8 (Appropriate): Emotion well-calibrated, shows natural engagement and authentic, spontaneous responses

- 9–10 (Highly Appropriate): Emotion nuanced, proportionate, context-sensitive, shows authentic human warmth, spontaneity, and genuine connection when appropriate

Important: Penalize responses that are overly formal, overly polite, template-like, or feel scripted, as these lack the natural spontaneity and emotional authenticity of real human interaction.
\\
Conversation:

VERSION A:
conversations a

VERSION B:
conversations b\\

Respond in JSON format:

"reasoning": "Analysis and comparison of emotional appropriateness between the two versions",

"version a score": 0-10,

"version b score": 0-10,

}
\end{AIbox}
\caption{Emotional Appropriateness Evaluation Prompt Continued}
\end{figure*}

%% file: table/comparision.tex
\section{Comparison}
Please refer to Table \ref{tab:appendix-baseline-vs-ours} for comparison between the baseline and our method across shared base models.

We used a mix of open-weight and closed-source models: Qwen3-235B (235B total, 22B active), Kimi-K2-0905 (1T total, 32B active), and MiMo-v2-Flash (309B total, 15B active) are open-weight MoE models, while GPT-5.2, GPT-4o-mini, and Grok-4-Fast are closed-source and do not have publicly disclosed parameter counts.

\begin{table*}[h] 
\centering 
\small 
\setlength{\tabcolsep}{4pt} 
\renewcommand{\arraystretch}{1.15} 
\resizebox{\linewidth}{!}{ 
\begin{tabular}{llcccccc} 
\toprule 
& & \multicolumn{3}{c}{\textbf{Emotional Intelligence}} & \multicolumn{3}{c}{\textbf{Social Competence}} \\ 
\cmidrule(lr){3-5} \cmidrule(lr){6-8} 
Evaluator & Model & Communication (COM) & Empathy (EMP) & Appropriateness (APP) & Continuity (CON) & Believability (BEL) & Social Rules (SOC) \\ 
\midrule 
\multicolumn{8}{l}{\textit{Baseline}} \\ 
\multirow{3}{*}{Sonnet-4.5} & GPT-4o-mini & 6.028 & 5.090 & 4.463 & 2.340 & 4.775 & 0.000 \\ 
& GPT-5.2 & 5.409 & 5.285 & 5.582 & 2.090 & 4.353 & -0.120 \\ 
& Grok-4 & 5.791 & 5.910 & 4.996 & 4.370 & 4.540 & 0.000 \\ 
\dashedrule
\addlinespace[0.5em]
\multirow{3}{*}{GPT-5.1} & GPT-4o-mini & 7.618 & 5.315 & 6.978 & 2.725 & 6.287 & 0.000 \\ 
& GPT-5.2 & 7.613 & 5.248 & 6.346 & 3.169 & 5.897 & 0.000 \\ 
& Grok-4 & 7.856 & 6.090 & 6.800 & 5.987 & 5.979 & 0.000 \\ 
\dashedrule
\addlinespace[0.5em]
\multirow{3}{*}{DeepSeek-V3.2} & GPT-4o-mini & 6.688 & 5.035 & 5.708 & 1.858 & 5.193 & 0.000 \\ 
& GPT-5.2 & 6.469 & 5.045 & 5.514 & 1.554 & 4.732 & 0.000 \\ 
& Grok-4 & 7.120 & 5.740 & 6.976 & 5.907 & 5.450 & 0.000 \\ 
\midrule 
\multicolumn{8}{l}{\textit{Ours}} \\ 
\multirow{3}{*}{Sonnet-4.5} & GPT-4o-mini & 6.973 & 6.387 & 5.059 & 3.938 & 3.540 & -0.023 \\ 
& GPT-5.2 & 9.201 & 7.072 & 8.799 & 6.732 & 8.051 & -0.581 \\ 
& Grok-4 & 8.574 & 5.696 & 7.348 & 5.666 & 7.285& -0.416 \\ 
\addlinespace[0.3em]
\dashedrule
\addlinespace[0.5em]
\multirow{3}{*}{GPT-5.1} & GPT-4o-mini & 7.903 & 6.418 & 6.700 & 5.378 & 4.924 & 0.000\\ 
& GPT-5.2 & 9.250 & 6.765 & 8.583 & 6.662 & 7.972 & 0.000 \\ 
& Grok-4 & 8.765 & 5.625 & 8.074 & 5.603 & 7.470 & -0.015 \\ 
\addlinespace[0.3em]
\dashedrule
\addlinespace[0.5em]
\multirow{3}{*}{DeepSeek-V3.2} & GPT-4o-mini & 8.363 & 5.953 & 7.138 & 5.377 & 4.843 & -0.043 \\ 
& GPT-5.2 & 9.573 & 6.170 & 8.838 & 6.458 & 8.593 & -0.122 \\ 
& Grok-4 & 8.925 & 5.415 & 8.355& 6.522 & 8.130 & -0.097 \\ 
\bottomrule 
\end{tabular}} 
\caption{Comparison between the baseline and our method across shared base models.}
\label{tab:appendix-baseline-vs-ours}
\end{table*}

%% file: appendix/network_metrics.tex
\section{Network Analysis Implementation}
\label{appendix:network-analysis}

This appendix provides complete implementation details for the network metrics reported in Table~2, ensuring reproducibility of our analysis.

\subsection{Graph Construction}

We constructed a directed, weighted social graph $G_t = (V, E_t, w_t)$ at each timestep $t$, where nodes represent agents and edge weights represent relationship strength. To focus on salient social ties, we applied an edge filtering threshold:

\begin{equation}
    E_t = \{(u, v) : w_{u \to v} \geq \tau\}, \quad \tau = 0.2
\end{equation}

This excludes negative weights and weak relationships (approximately 5--10\% of edges), following standard practice in weighted network analysis \citep{newman2004analysis}.

\subsection{Network Metrics}

\paragraph{Modularity ($Q$).}
We applied the Louvain community detection algorithm \citep{blondel2008fast} to detect emergent community structure. Since Louvain operates on undirected graphs, we symmetrized the directed graph by summing bidirectional edges: $w_{\{u,v\}} = w_{u \to v} + w_{v \to u}$. Modularity was computed as:

\begin{equation}
    Q = \frac{1}{2W} \sum_{i,j} \left[ w_{ij} - \frac{s_i s_j}{2W} \right] \delta(c_i, c_j)
\end{equation}

\noindent where $s_i = \sum_j w_{ij}$ is the weighted degree, $W = \sum_{ij} w_{ij}$ is the total edge weight, and $\delta(c_i, c_j) = 1$ if nodes $i$ and $j$ belong to the same community. We used the default resolution parameter $\gamma = 1.0$ and fixed \texttt{random\_state=42} for reproducibility.

\paragraph{Reciprocity ($r$).}
We measured the fraction of node pairs with bidirectional edges in the directed graph:

\begin{equation}
    r = \frac{|\{(u,v) : (u,v) \in E \land (v,u) \in E\}|}{|\{(u,v) : (u,v) \in E \lor (v,u) \in E\}|}
\end{equation}

\paragraph{Weighted Reciprocity ($r_w$).}
For reciprocal pairs, we computed weight similarity as:

\begin{equation}
    r_w = \frac{1}{|R|} \sum_{(u,v) \in R} \left(1 - \frac{|w_{u \to v} - w_{v \to u}|}{w_{u \to v} + w_{v \to u}}\right)
\end{equation}

\noindent where $R$ denotes the set of reciprocal pairs. Values range from 0 (asymmetric) to 1 (perfectly symmetric).

\paragraph{Normalized Mutual Information (NMI).}
To measure community stability across timesteps, we computed NMI between consecutive partitions \citep{strehl2003cluster}:

\begin{equation}
    \text{NMI}(C_t, C_{t+1}) = \frac{2 \cdot I(C_t; C_{t+1})}{H(C_t) + H(C_{t+1})}
\end{equation}

\noindent where $I(\cdot;\cdot)$ is mutual information and $H(\cdot)$ is entropy. Only nodes present at both timesteps were compared, following standard practice in dynamic community detection \citep{rossetti2018community}.

\paragraph{Weighted Drift.}
We measured the fraction of nodes changing community membership, weighted by node importance (total edge weight):

\begin{equation}
    \text{Drift}_w = \frac{\sum_{v : c_v^{(t)} \neq c_v^{(t+1)}} s_v^{(t)}}{\sum_{v \in V_{\text{common}}} s_v^{(t)}}
\end{equation}

\noindent This weights high-degree hub transitions more heavily than peripheral node changes.

\subsection{Edge Case Handling}

\paragraph{Negative Weights.} Edges with negative relationship strength were excluded, as standard modularity is not well-defined for signed graphs without specialized algorithms \citep{yang2007community}.

\paragraph{Isolated Nodes.} Nodes with no edges above threshold $\tau$ were assigned to singleton communities by Louvain and excluded from NMI/Drift calculations.

\paragraph{Graph Symmetrization.} For modularity computation only, we summed bidirectional edges rather than averaging, preserving total relationship strength between node pairs \citep{borgatti2009network}.

\subsection{Validation}

We validated our implementation with synthetic tests:

\begin{itemize}
    \item \textbf{Perfect stability}: 25 agents in 3 fixed communities across 10 timesteps yielded $Q = 0.797$, NMI $= 1.0$, Drift $= 0.0$ (expected).
    \item \textbf{Complete rewiring}: Random edge reassignment at $t=5$ yielded NMI $= 0.021$, Drift $= 0.983$ (expected $\approx 0$ and $\approx 1$).
\end{itemize}

\subsection{Software}

All analyses used Python 3.10 with \texttt{python-louvain} v0.16, \texttt{scikit-learn} v1.3.0, and \texttt{networkx} v3.1.